\documentclass[twoside]{article}

\usepackage{microtype}
\usepackage{graphicx}
\usepackage{subfigure}
\usepackage{float}
\usepackage{lipsum}
\usepackage{booktabs} % for professional tables
\usepackage{nicefrac}       % compact symbols for 1/2, etc.
\usepackage{adjustbox}

\usepackage{hyperref}
\usepackage{wrapfig}
\usepackage[authoryear, round]{natbib}

\newcommand{\name}{MineGrad}
\newcommand{\namespace}{MineGrad }

\usepackage[accepted]{aistats2026}

\usepackage{amsmath}
\usepackage{amssymb}
\usepackage{mathtools}
\usepackage{amsthm}
\usepackage{afterpage}
\usepackage[capitalize,noabbrev]{cleveref}

\theoremstyle{plain}

\theoremstyle{definition}

\theoremstyle{remark}

\usepackage[textsize=tiny]{todonotes}
\usepackage{multirow} 
\usepackage{multicol}
\usepackage{subcaption}
\usepackage{graphicx}
\usepackage{longtable}
\usepackage{amsmath} 
\allowdisplaybreaks

\title{\name: Gradient Inversion Attacks on LoRA Fine-Tuning}


\begin{document}

\twocolumn[
\aistatstitle{MineGrad: Gradient Inversion Attacks on LoRA Fine-Tuning}

\aistatsauthor{Hasin Us Sami\textsuperscript{*}, Swapneel Sen\textsuperscript{*}, Ba\c{s}ak G\"{u}ler}
\aistatsaddress{University of California, Riverside, CA\\
{\tt\small hsami003@ucr.edu, ssen010@ucr.edu, bguler@ece.ucr.edu}
}
]

\footnotetext[1]{Equal contribution.}

\begin{abstract}
Parameter-efficient fine-tuning (PEFT), such as low-rank adaptation (LoRA), has recently been adopted in federated learning  to reduce communication and computation costs. In this setup, users download a pretrained model from the server prior to fine-tuning, and then fine-tune lightweight LoRA modules locally while keeping the pretrained model frozen, sharing only the gradients of the fine-tuning parameters with the server. 
Despite its growing popularity, robustness of federated fine-tuning against an adversarial server remains underexplored,
where the server maliciously tampers with the training protocol
to breach the privacy of users’ data.
In this work, we investigate gradient inversion attacks on LoRA fine-tuning.  We propose an analytical attack that enables a malicious server to recover private user data by leveraging a poisoned pretrained model and fine-tuning parameters. Our design embeds fine-tuning data within the shared gradients, to allow the server to analytically reconstruct user data. Unlike prior works, our attack is applicable to both language and vision tasks, does not rely on computationally expensive (adversarial) pretraining with public datasets or require the number of training tokens to be less than the rank of LoRA modules. 
Experimental results on both language and vision tasks demonstrate high-fidelity data recovery across multiple baselines, revealing several critical vulnerabilities.

\end{abstract}

\section{INTRODUCTION}\label{sec:intro}

Large-scale pretrained models achieved remarkable success across various tasks. However, their immense size—often billions of parameters \cite{Radford2019, Jacob2019, Yinhan2019, Brown2020, He2021, Hugo2023}—renders full fine-tuning (FFT) computationally and storage intensive. Parameter-efficient fine-tuning (PEFT) mitigates this by updating only a small subset of parameters, either integrated into the pretrained model \cite{Han2020, Bingchen2024, Kowsher2024} or added as external modules \cite{Houlsby2019, Rabeeh2021, Edward2022, Hayou2024}.

To reduce communication, computation and memory overhead, PEFT is increasingly adopted in federated learning (FL) \cite{Zhang2023, Yeachan2023, Shysheya2023, Yae2024}, where users collaboratively train a shared model without revealing local data. While FL’s core principle is on-device privacy, shared gradients remain vulnerable. Gradient inversion attacks can reconstruct training data \cite{Zhu2019, Geiping2020, Yin2021, Yangsibo2021, Lu2022, Deng2021, Gupta2022, Fowl2023}, while membership inference \cite{Shokri2017, Nasr2019, Song2021, Fatemeh2022, Francisco2023} and data extraction \cite{Carlini2018, Carlini2021, Florian2022, Panda2024} attacks expose additional risks. Among these, gradient inversion is especially dangerous since it requires no prior knowledge of target samples.

Despite these concerns, gradient inversion attacks on PEFT  remain largely underexplored. 
 \cite{Feng2024} recently introduced a gradient inversion attack by deploying a poisoned pretrained model, focusing on the FFT setup, where the attacker has access to gradients for the full model parameters. \cite{Liu2024} and \cite{Wen2024} explore manipulating the pretrained model for membership inference and data extraction attacks in FFT and PEFT settings. 
Recent observations suggest that PEFT methods may offer better privacy protections against inversion attacks by reducing the number of leaked parameters \cite{Zhang2023}.  
However, this setup does not consider stronger adversaries, such as an adversarial server that can manipulate the pretrained model along with shared PEFT parameters before sending it to users \cite{Fowl2022, Fowl2023, LOKI2024, Feng2024}. In such cases, a target user (victim) might inadvertently download a tampered pretrained model and fine-tuning parameters from a compromised server.

This work investigates gradient inversion attacks on low-rank adaptation (LoRA) \cite{Edward2022}, where users fine-tune small low-rank matrices while keeping the pretrained model frozen \cite{Zhang2023, Sun2024, Bian2024, Yae2024}. This reduced gradient space complicates inversion. While recent work has explored privacy-aware LoRA  \cite{Da2022, Sun2024, Xianzhi2024}, practical gradient inversion attacks remain underexplored, for instance, whether adversarial servers can invert LoRA gradients for vision transformers (ViT) \cite{Dosovitskiy2021}. 
Our work answers these questions in the affirmative.

Recently, \cite{Dixi2024} considered attacks on LoRA-based fine-tuning of a pretrained diffusion model to generate private fine-tuning data. The attack leverages prior training of a neural network encoder using a publicly available dataset from a similar domain to the private fine-tuning data. In contrast, here we do not assume availability of publicly available datasets from the fine-tuning domain - as such data may not be available in practice, especially in privacy-sensitive settings such as FL. \cite{gao2025gradient} proposes a discrete optimization attack, which requires access to the gradients of the word embedding layer. In LoRA fine-tuning, however, word embedding layer is part of the frozen pretrained model and the server loses access to this gradient, hence the attack cannot be applied to our problem.

Analytical attacks are proposed to recover fine-tuning data from gradients in the context of FFT \cite{Feng2024} or PEFT with adapters \cite{sami2025gradient} of a pretrained model. These works utilize a carefully crafted pretrained model to facilitate undistorted propagation of target token or patch embeddings through the network up to the trainable linear layer, followed by a non-linear activation function like GELU or ReLU. Subsequently, the weight and bias parameters are precisely designed to ensure that no two tokens or patches activate the same set of neurons. Thus, the corresponding weight and bias gradients can be utilized to accurately recover the original data.
In contrast, in LoRA fine-tuning, trainable low-rank matrices are typically not followed by such an activation function, making the aforementioned attacks infeasible.

Another notable work is \cite{Petrov2024}, which proposed a gradient inversion attack, DAGER, to recover text data, by leveraging the low-rank structure of weight gradients. This attack can also be applied to LoRA fine-tuning. However, the attack requires the total number of training tokens in the batch to be less than or equal to the rank of LoRA matrices, which is often unrealistic in practice \cite{Edward2022}. 
In addition, DAGER relies on an exhaustive search across all the tokens in the vocabulary in all possible positions, hence it cannot be applied to datasets that do not have a predefined vocabulary such as vision datasets.

To address these challenges, we introduce a novel gradient inversion attack, \name,  %on LoRA fine-tuning, 
by leveraging the gradients corresponding to the trainable low-rank matrices. We design a poisoned pretrained model and low-rank matrices to embed the fine-tuning data within the victim’s gradients. Using these manipulated gradients, the server can analytically extract the fine-tuning data. Unlike prior methods, our attack can be applied to both language and vision tasks, does not rely on computationally expensive pretraining with a public dataset, and does not require that the number of tokens be less than the rank of the LoRA matrices \cite{Petrov2024}. Our code is available at 
\url{https://github.com/info-ucr/MineGrad}.

\section{RELATED WORKS} \label{sec:related_works}

\emph{Parameter-Efficient Fine Tuning and LoRA. } PEFT mechanisms reduce the storage and computation costs of fine-tuning large pretrained models by freezing the model and updating lightweight modules \cite{Houlsby2019, Han2020, Rabeeh2021, Lisa2021, Edward2022}. 
%Among all PEFT mechanisms, 
Among the PEFT techniques, LoRA and its variants are among the most widely-adopted due to their superior performance in resource-limited settings \cite{Edward2022, Qingru2023, Yeming2024, Hayou2024}. Recent works have further extended the application of LoRA to FL \cite{Zhang2023, Yae2024, jianhao2024}.

\emph{Gradient inversion attacks. } Gradient inversion attacks aim to reconstruct training data from shared gradients. To do so, earlier works retrieve data by solving a distance minimization problem \cite{Zhu2019, Geiping2020} or training a generative adversarial network \cite{Hitaj2017, Zhibo2019, Yuheng2020}. More recent works leverage batch normalization statistics \cite{Yin2021, Ali2022}, dropout mask optimization \cite{Daniel2023} or blind source separation \cite{Sanjay2023} to enhance the attack performance. Beyond attacks to recover image samples, several works propose mechanisms to recover text or tabular data \cite{Deng2021, Gupta2022, Mislav2022, DVero2023}. Attack performance can further be enhanced by considering stronger adversaries who can tamper with model parameters or architectures \cite{Fowl2022, Hong2023, Fowl2023, Feng2024}. A gradient-inversion attack was recently demonstrated for PEFT with adapters \cite{sami2025gradient}. To recover data, this work leverages weight and bias gradients from adapter modules, along with distinct neuron activations within a linear layer. 
Other lines of work consider membership inference and data extraction attacks \cite{Shokri2017, Nasr2019, Song2021, Fatemeh2022, Francisco2023, Carlini2021,Liu2024}.
\section{PRELIMINARIES AND PROBLEM SETUP} \label{sec:prob_for}

% ====== PREAMBLE (put before \begin{document}) ======
%\usepackage{wrapfig}
%\usepackage{placeins}   % for \FloatBarrier if needed
\setlength{\intextsep}{8pt}   % space around figures in text
\setlength{\columnsep}{7pt}   % space between text and wrapfigure

% ====== BODY ======

\textbf{LoRA fine-tuning. }
In a typical transformer~\cite{Vaswani2017, Jacob2019, Yinhan2019}, a distinct
position encoding vector is added to each token. These token
embeddings are then passed through transformer encoders, where each encoder
consists of stacked multi-head self-attention (MSA), multi-layer perceptron
(MLP) and LayerNorm (LN) layers. 
\begin{figure}[ht]
  \centering
  \includegraphics[width=0.85\linewidth]{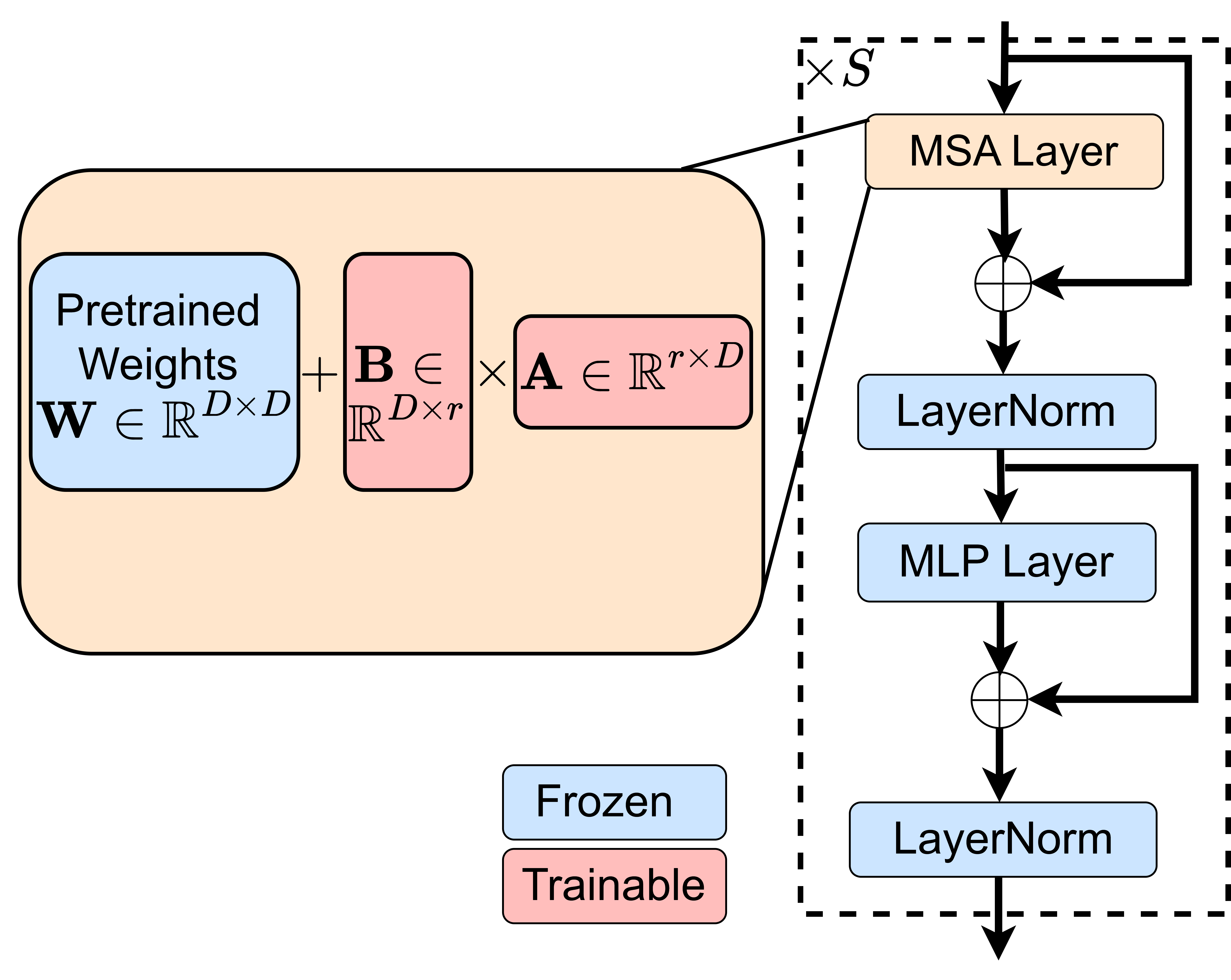} % slightly smaller
  \caption{{\bf Transformer encoder with LoRA modules.} Transformer encoder consists of stacked MSA, LayerNorm, MLP layers and residual connections. Low-rank matrices $\mathbf{A}, \mathbf{B}$ (of rank $r$) are inserted within each MSA layer.}
  \label{fig:Roberta}
\end{figure}
For LoRA fine-tuning, low-rank decomposition matrices $\mathbf{A}$, $\mathbf{B}$
(of rank $r$) are used as additive modules within each MSA layer as illustrated
in Fig.~\ref{fig:Roberta}. The product $\mathbf{BA}$ is added to the pretrained
MSA weights during inference. During backpropagation, only the gradients with
respect to $\mathbf{A}$ and $\mathbf{B}$ are computed, while keeping the
pretrained weights frozen.

\textbf{Federated LoRA fine-tuning.} 
We consider a centralized FL setting with $U$ users. For LoRA-based fine-tuning,
we denote the low-rank matrices of user $u \in [U]$ as
$\mathbf{W}_F^u \triangleq \{\mathbf{A}^u, \mathbf{B}^u\}$. Prior to training,
the server shares a pretrained model $\mathbf{W}_P$ with the users. In each
training round, user $u \in [U]$ performs fine-tuning on its local dataset and
sends a local update (gradient) $\Delta \mathbf{W}_F^u$ to the
server~\cite{jianhao2024}. After receiving the local updates, the server
computes the current state of the global model
$\mathbf{W}_F \triangleq \{\mathbf{A}, \mathbf{B}\}$:
\begin{align} \label{eq:FL}
    \mathbf{W}_F = \mathbf{W}_F + \frac{1}{U}\sum_{u=1}^U \Delta \mathbf{W}_F^u
\end{align}
The goal is to minimize the global loss function,
$\mathcal{L} \triangleq \frac{1}{U}\sum_{u=1}^U
\mathcal{L}_u(\mathbf{W}_P, \mathbf{W}_F)$, where $\mathcal{L}_u$ denotes the
local loss of user $u$ and $\mathbf{W}_P$ denotes the frozen pretrained model.

\begin{figure}[t]
  \centering
  \includegraphics[width=0.85\columnwidth]{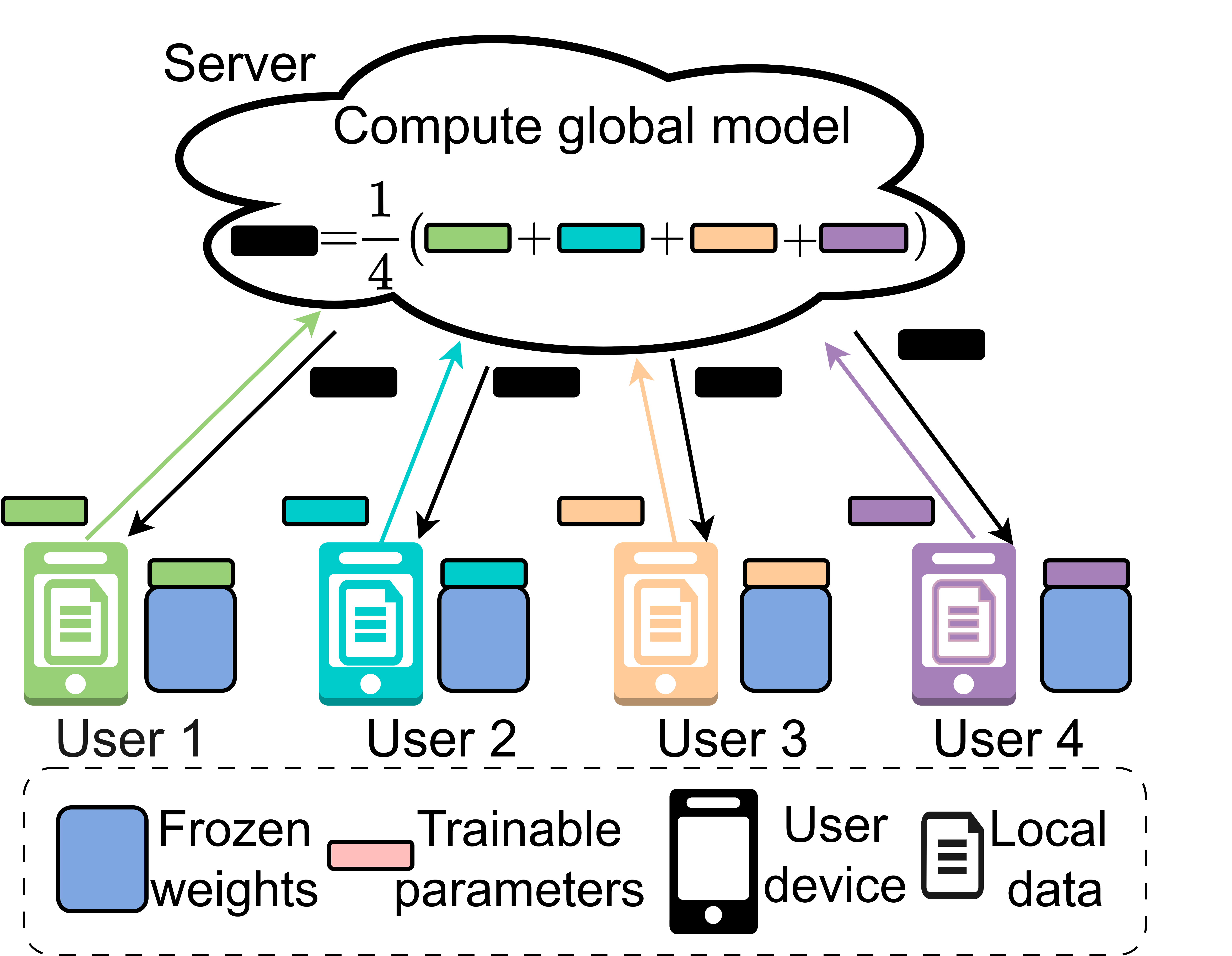}
  \caption{{\bf Federated LoRA with $U=4$ users.} In each training round, users perform fine-tuning locally and send the LoRA gradients to the server. The server performs aggregation and sends the new global state of the LoRA modules to the users.}
  \label{fig:FL}
\end{figure}

{\bf Threat model.} We consider an adversarial server who can tamper with the model parameters shared with users. This includes: 1) disseminating a poisonous pretrained model to users once prior to fine-tuning, and 2)  modifying the  global fine-tuning parameters sent to the users in each training round after aggregation. As in \cite{John2023, Shysheya2023, Chen2023}, the server performs pretraining with proprietary data or publicly available proxy datasets and then releases the pretrained model to the edge users. Hence, the server can manipulate the pretrained model (once before training) and global fine-tuning parameters (at each round during training).  The adversary aims to coerce a target user into producing gradients with respect to the poisonous fine-tuning modules, to reveal its local data. Malicious tampering with the training protocol is inspired by \cite{Fowl2022, Fowl2023, Hong2023, LOKI2024, Feng2024}.
Users often download a pretrained model without any formal verification protocol \cite{Feng2024, Wen2024, Liu2024}. Training is usually performed in the background e.g., during night time when the device is being charged or remains idle \cite{Ramaswamy2019, Fowl2023}. Therefore, a user may not be observing the training loss on a continual basis. Hence, even though a malicious pretrained model may adversely affect the training performance, a user may fail to detect such suspicious activity under the common practices of FL.

{\bf Existing Attacks.} 
%%%%%%%%%%%%%%%%%%%%%%%%%%%%%%%%%%
The closest to our work is DAGER \cite{Petrov2024}, which leverages rank-deficiency of weight gradients to recover the input tokens. Let us denote the total number of input tokens as $N$, where each token embedding is of dimension $D$. For a linear layer with stacked input tokens $\mathbf{Y} \in \mathbb{R}^{N \times D}$, weight $\mathbf{W} \in \mathbb{R}^{D \times r}$, and bias $\mathbf{b} \in \mathbb{R}^{r}$, the output can be written as $ \mathbf{Z} \triangleq \mathbf{Y}\mathbf{W}+ (\mathbf{b}|\ldots |\mathbf{b})^{\text{T}}$. Accordingly,
 the gradient of user $u$ can be written as,
%the loss function $\mathcal{L}_u$ (of user $u$) to be
%\vspace{-0.2cm}
\begin{align}
\frac{\partial \mathcal{L}_u}{\partial \mathbf{W}} \triangleq \mathbf{Y}^{\text{T}}\frac{\partial \mathcal{L}_u}{\partial \mathbf{Z}}
\end{align}
The rank of $\frac{\partial \mathcal{L}_u}{\partial \mathbf{W}}$ is at most $N$ if $N \leq min\{D,r\}$. Due to the rank-deficiency of the gradient matrix $\frac{\partial \mathcal{L}_u}{\partial \mathbf{W}}$, its columns form a subspace of $\mathbb{R}^D$ with dimension $N$.  Next, DAGER initiates an exhaustive search algorithm with all possible tokens in the vocabulary in all possible positions. The goal is to find the set of target input embeddings by checking whether they are in the column space of $\frac{\partial \mathcal{L}_u}{\partial \mathbf{W}}$. Overall, the success of the reconstruction algorithm relies on the assumption that the total number of training tokens $N$ satisfies: $N \leq min\{D,r\}$.

{\bf Challenges. } 
With LoRA  fine-tuning, the attacker has a much reduced space of observable gradients. % making token reconstruction considerably harder. 
The attacker only has access to $r$ gradient vectors per low-rank matrix in each MSA layer. As noted in \cite{Edward2022}, the value of $r$ typically ranges from $1$ to $64$. 
When applying DAGER \cite{Petrov2024} to LoRA fine-tuning, the attack success requires the total number of training tokens to be less than or equal to the rank, $r$, of the LoRA matrices. This condition is often not satisfied in practice since the rank of the LoRA matrices is typically small as noted above. 
In addition, DAGER relies on an exhaustive search with all possible words in the vocabulary in all possible positions, which is not feasible for datasets without a predefined vocabulary such as vision datasets. 

\begin{figure*}[t]
  \centering
   \includegraphics[width=1.03\linewidth, trim=0.4cm 0cm 0.5cm 0cm, clip]{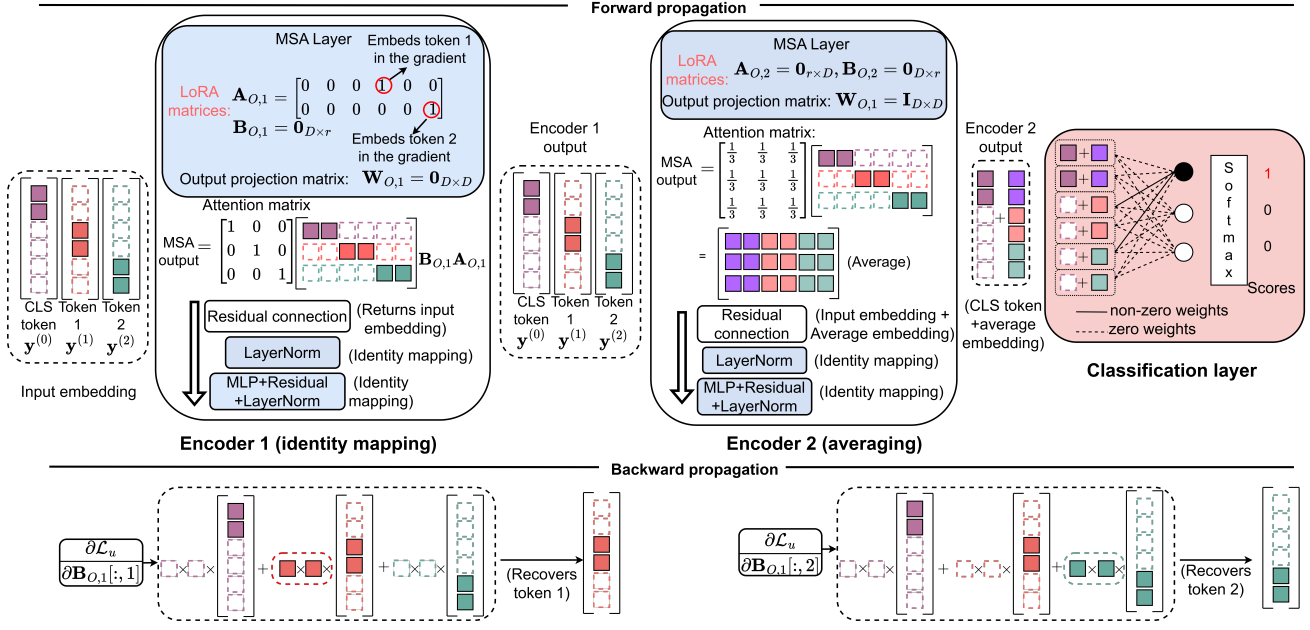}
   %\vspace{-1cm}
   \vspace{-0.5cm} \caption{ Overview of \namespace with $D=6$, $r=2$, $N=2$. Each token embedding $\mathbf{y}^{(n)}$ for $n=0,1,2$ contains large values in two distinct positions (colored). The $1$st encoder is designed to output identity mapping of the input tokens. 
   The LoRA matrices $\mathbf{A}_{O,1}$, $\mathbf{B}_{O,1}$ in the $1$st encoder are designed to recover tokens $1$ and $2$. 
   The $2$nd encoder is designed to produce average of the tokens. In the classification layer, only the embedding for the class token is utilized. Only the $3$rd and $5$th elements of this embedding are connected to the $1$st logit via large weights, leading the $1$st logit to produce a softmax score of $1$. In the gradient for the $1$st column of $\mathbf{B}_{O,1}$, each embedding is weighted by $\mathbf{y}^{(i)}_3 \mathbf{y}^{(i)}_4$ for token $i$, where $\mathbf{y}_d^{(i)}$ is the $d^{th}$ element of $\mathbf{y}^{(i)}$. Token $1$ has a larger weight compared to the other tokens, which enables the recovery of token $1$. In the gradient for the $2$nd column of $\mathbf{B}_{O,1}$, each embedding is weighted by $\mathbf{y}^{(i)}_5 \mathbf{y}^{(i)}_6$ for token $i$, leading to the  recovery of token $2$. 
      } 
   \label{fig:design}
\vspace{-0.4cm}\end{figure*}

{\bf Contributions. } To address these challenges, we propose a novel inversion attack, \name, for LoRA fine-tuning in FL. Our approach involves a carefully crafted design of the pretrained model, which remains fixed during fine-tuning but is utilized during inference. Additionally, we demonstrate an adversarial manipulation of the lightweight LoRA matrices, which are shared with the users in each training round during federated fine-tuning. These two adversarial components work in tandem to reveal fine-tuning data of a target user effectively. To maximize the number of tokens reconstructed, \namespace mitigates the low-rank issue of a single matrix by utilizing the  gradients from multiple LoRA matrices across different encoders.  Unlike \cite{Petrov2024}, \namespace can successfully recover data even when the number of training tokens are greater than the rank of LoRA matrices and can be applied to vision datasets.

\section{MINEGRAD} \label{sec:framework}
For simplicity,  we initially consider text classification as the downstream task, and provide the key design principles of \name. Details are deferred to  Apps.~\ref{App:1} and \ref{App:2}. 
 For the pretrained model, we first consider encoder-based transformer architectures building on bidirectional self-attention, such as  BERT \cite{Jacob2019} and its successor  RoBERTa \cite{Yinhan2019}, which has been used to demonstrate the efficacy of LoRA over other PEFT methods \cite{Edward2022}. Our framework can also be applied to decoder-based architectures with unidirectional self-attention such as GPT-2 \cite{Radford2019}, as we demonstrate in App. \ref{app:GPT2}. In our experiments, we also extend our results to vision transformers (ViT) for image classification.

Fig.~\ref{fig:Roberta} illustrates the transformer encoder with LoRA modules. 
The typical layers within an encoder, i.e., MSA, LN, and MLP are frozen, while only the LoRA matrices and classification head are trainable. 
For simplicity, here we consider fine-tuning on a single text sequence, consisting of $N$ tokens.
Our framework can also handle a batch of sequences, as shown in Section \ref{sec:exp} along with the theoretical analysis in App. \ref{App:4}.

Denote the word embedding of token $n\in[N]$ as $\mathbf{x}^{(n)} \in \mathbb{R}^D$, and the class token as $\mathbf{x}^{(0)}$. 
Embedding $\mathbf{x}^{(n)}$ is  added to a  position encoding vector  $\mathbf{e}^{(n)} \in \mathbb{R}^D$, resulting in a token embedding, 
\vspace{-0.1cm}
\begin{align} \label{eq:y}
    \mathbf{y}^{(n)} \triangleq \mathbf{x}^{(n)}+\mathbf{e}^{(n)}  \text{ for } n \in\{0, \ldots, N\}
\end{align}
We now demonstrate a malicious parameterization of the pretrained model and global LoRA modules for the server to recover the private fine-tuning  tokens for a set of target positions $\mathcal{T}$, using the gradients shared by a victim user $u$. 

%\vspace{-0.1cm}
{\bf Key intuition. } \namespace   is an adversarial design where each column of the low-rank adaptation matrix $\mathbf{B}$ reveals a token at a (different) targeted position via its gradient. Since $\mathbf{B}$ has rank $r$, at most $r$ tokens can be revealed per encoder. To scale, we employ multiple encoders and carefully structure self-attention and low-rank matrices $\mathbf{A}$ and $\mathbf{B}$ to preserve token embeddings via identity mappings in the forward pass. During backpropagation, each encoder then exposes private tokens from distinct target positions.  
Our design ensures that for each column $n$ of the fine-tuning matrix $\mathbf{B}$,  the local gradient shared by the user contains a weighted average $\sum_{i\in\mathcal{T}} \alpha_{n, i} \mathbf{y}^{(i)}$ of the token embeddings $\mathbf{y}^{(i)}$, where the weight $\alpha_{n,i}$ is much larger for the target token compared to all other tokens.  %$i\neq n$,  
% (i.e., $\alpha_{n,n}\gg \alpha_{n,i}$ for all $i\neq n$), 
This enables recovery of a distinct  token embedding from each gradient. The weighting is controlled by the position encoding vector $\mathbf{e}^{(n)}$. We illustrate this idea in Fig.~\ref{fig:design}.

\noindent
{\bf Position Encoding. }
Since position encoding vectors are part of the pretrained model, the attacker has the capability to tamper with these parameters once \cite{Feng2024}.
 Let $H$ be the total number of heads in the MSA layer  and $\overline{D}$ be the dimension of each head, i.e., $\overline{D} \triangleq D/H$. 
The server designs the position encoding vectors as,
\vspace{-0.1cm}
\begin{align} \label{eq:Epos}
 \mathbf{e}_{d}^{(n)}\hspace{-0.1cm}\triangleq \hspace{-0.1cm}\left\{\begin{matrix} 
 c_1 &\text{ if } d=2n+1 \\
 -c_1 &\text{ if } d=2n+2\\
 c_2 &\hspace{-0.2cm}\text{ if } d=2n+1 + (h-1)\overline{D} \text{ for } 2\leq h\leq H\\
 -c_2 &\hspace{-0.2cm}\text{ if } d=2n+2+ (h-1)\overline{D} \text{ for } 2\leq h\leq H\\
 0 &\text{ otherwise }
 \end{matrix} \right. \vspace{-0.2cm}
\end{align}
where $\mathbf{e}_{d}^{(n)}$ is the $d^{th}$ element of $\mathbf{e}^{(n)}$, and $c_1$ is a large positive value ($\sim10^2$) such that $c_1>>c_2$. We want $c_2$ to have negligible effect on the standard deviation of $\mathbf{e}^{(n)}$ compared to $c_1$,  while ensuring, 
\vspace{-0.1cm}
\begin{equation} \label{eq:condn1}
(\mathbf{e}^{(i)})_h^{\text{T}} (\mathbf{e}^{(i)})_h >> (\mathbf{e}^{(i)})_h^{\text{T}} (\mathbf{e}^{(j)})_h \quad \text{ if } i\neq j
\end{equation}
% for $i \neq j$ where
for all heads $h \in [H]$, where 
 $(\mathbf{e}^{(i)})_h \triangleq  [
 \mathbf{e}_{h\overline{D}+1}^{(i)} \cdots  \mathbf{e}_{(h+1)\overline{D}}^{(i)}
]$
represents the $\overline{D}$ elements that propagate through head $h$.  As the word embeddings $\mathbf{x}^{(n)}$ typically lie within the range $[-1,1]$ \cite{wolf2020}, $\mathbf{e}^{(n)}$ is the dominant factor in determining the mean and standard deviation across the elements in $\mathbf{y}^{(n)}$ from \eqref{eq:y}.  
From \eqref{eq:Epos}, the mean across the elements in $\mathbf{y}^{(n)}$ is,
\vspace{-0.2cm}
\begin{align} \label{eq:meanY}
    \mu^{(n)}\triangleq \frac{1}{D}\sum_{d=1}^D \mathbf{y}^{(n)}_d \approx \frac{1}{D}\sum_{i=1}^D \mathbf{e}_{d}^{(n)}=0 \triangleq \mu
\end{align}

whereas the standard deviation is,
\vspace{-0.1cm}
\begin{align} \label{eq:stdY}
    \sigma^{(n)} \approx \sqrt{\frac{1}{D}\sum_{d=1}^D (\mathbf{e}^{(n)}_d-\mu^{(n)})^2}  \triangleq \sigma   
\end{align}
\begin{figure}[ht]
  \centering
  \includegraphics[width=0.85\linewidth]{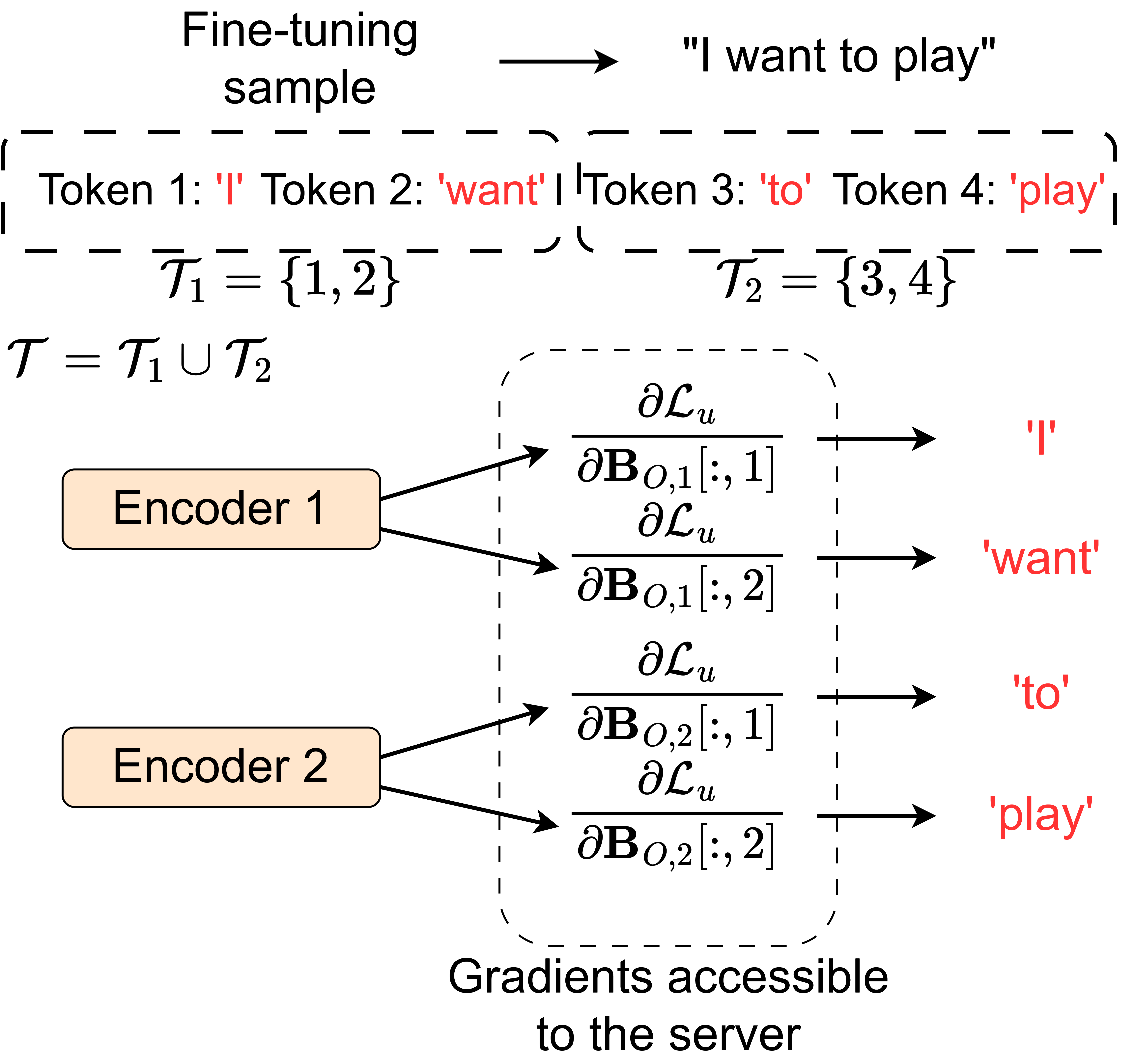}
  \vspace{-0.2cm} 
  \caption{
   {\bf Token extraction from multiple encoders.} The rank of the LoRA modules is $r=2$. 
   Gradients from Encoder $1$'s LoRA modules recover tokens $1$ and $2$ in the target set $\mathcal{T}_1=\{1,2\}$.  Gradients from Encoder $2$'s LoRA modules recover tokens $3$ and $4$, i.e.,  $\mathcal{T}_2=\{3,4\}$. 
  } 
  \label{fig:attack}
  \end{figure}
  
As we describe later, this design will be critical in ensuring that each target token can be recovered from the accumulated gradients for  the LoRA modules.

\vspace{-0.1cm}
\noindent
\textbf{Target Encoders (Encoders $1$ to $S-1$). } 
Let  $S$ be the total number of encoders. 
We call encoders $1$ to $S-1$ the \emph{target encoders}, as the fine-tuning modules of these encoders will produce the gradients that reveal the target tokens. 
The set of tokens targeted by encoder $t\in[S-1]$ is denoted by $\mathcal{T}_t$. We denote the $n^{th}$ element in set $\mathcal{T}_t$ by $\mathcal{T}_t(n)$. LoRA modules are added to query, key, value, and output projection matrices, denoted by $\mathbf{W}_{Q,t}, \mathbf{W}_{K,t}, \mathbf{W}_{V,t}$ and $\mathbf{W}_{O,t}$ \cite{Edward2022}. Our attack recovers tokens from the gradients of LoRA modules added to $\mathbf{W}_{V,t}$ and $\mathbf{W}_{O,t}$. For simplicity, we describe our attack using $\mathbf{W}_{O,t}$. Same principles also apply to $\mathbf{W}_{V,t}$.

%{\it \underline{MSA layer:} } 
Embeddings $\mathbf{y}^{(n)}$ first enter the MSA layer.   
Our design  of query, key, value weights, biases, and LoRA matrices produces an attention matrix $\mathbf{I}_{(N+1) \times (N+1)}$ for all heads (App.~\ref{App:1}). 
Then, the self-attention output for head $h \in [H]$  becomes an identity mapping of the input. 
We set the output projection matrix as $\mathbf{W}_{O,t} = \mathbf{0}_{D \times D}$. After the trainable LoRA modules $\mathbf{A}_{O,t} \in \mathbb{R}^{r \times D}$, $\mathbf{B}_{O,t} \in \mathbb{R}^{D \times r}$ are inserted to $\mathbf{W}_{O,t}$,   
the MSA output becomes,
\begin{align}
&\text{MSA}([\mathbf{y}^{(0)}, \dots, \mathbf{y}^{(N)}]) \notag \\
&\triangleq 
\hspace{-0.0cm}\big[ SA_1(\mathbf{y}^{(0)}, \dots, \mathbf{y}^{(N)}), 
\dots, 
SA_H(\mathbf{y}^{(0)}, \dots, \mathbf{y}^{(N)}) \big] \notag \\
&\quad\quad\quad (\mathbf{W}_{O,t} + \mathbf{B}_{O,t}\mathbf{A}_{O,t}) \notag \\
&=
\begin{bmatrix}
\mathbf{y}^{(0)} & \dots & \mathbf{y}^{(N)}
\end{bmatrix}^{\!\!\text{T}}
\mathbf{B}_{O,t}\mathbf{A}_{O,t}
\label{eq:MSAop}
\end{align}

The LoRA modules are then designed as,
\vspace{-0.1cm}
\begin{align} \label{eq:design_A}
\mathbf{A}_{O,t}[p,q] &\triangleq 
\begin{cases}
1 & \text{if } p=n \text{ and } q=2\mathcal{T}_t(n)+2, \; n \in [r] \\
0 & \text{otherwise}
\end{cases} \notag \\
\mathbf{B}_{O,t} & = \mathbf{0}_{D \times r}
\end{align}
where $\mathbf{A}_{O,t}[p,q]$  is the element at row $p$, column $q$.
 The design in \eqref{eq:design_A} ensures that for any token $i$,
 \vspace{-0.1cm}
\begin{align}
    \frac{\partial \mathbf{A}_{O,t}^{\text{T}}[d,:]\mathbf{B}_{O,t}^{\text{T}} \mathbf{y}^{(i)}}{\partial \mathbf{B}_{O,t}[:,n]} 
    &=\left\{\begin{matrix}
    \mathbf{y}^{(i)} &\text{ if } d=2\mathcal{T}_t(n)+2\\
    \mathbf{0} &\text{ otherwise }
    \end{matrix} \right. \label{eq:der_ATBT}
\end{align} 
where $\mathbf{A}_{O,t}^\text{T}[d,:]$ is the $d^{th}$ row of $\mathbf{A}_{O,t}^\text{T}$, and $\mathbf{B}_{O,t}^\text{T}[:,n]$ is the $n^{th}$ column of $\mathbf{B}_{O,t}$. 
 As shown later, this enables the attacker to recover the $n^{th}$ token in $\mathcal{T}_t$ using the gradient with respect to $\mathbf{B}_{O,t}^\text{T}[:,n]$. Since $|\mathcal{T}_t| \leq r$, from a single encoder, at most $r$ tokens can be recovered. 
 % (from the gradients with respect to $r$ columns). 
 From \eqref{eq:design_A}, $\mathbf{B}_{O,t}\mathbf{A}_{O,t}=\mathbf{0}_{D \times D}$, hence,  
 \vspace{-0.1cm}
 \begin{align} \label{eq:AB=0}
    \begin{bmatrix}
         \mathbf{y}^{(0)} & 
        \cdots&
        \mathbf{y}^{(N)}
    \end{bmatrix}^{\text{T}}\mathbf{B}_{O,t}\mathbf{A}_{O,t}=\mathbf{0}
\end{align} 
and the output of the residual connection becomes,
\vspace{-0.05cm}
\begin{align}
\mathbf{u}^{(n)} \triangleq \mathbf{y}^{(n)}+\mathbf{A}_{O,t}^\text{T}\mathbf{B}_{O,t}^{\text{T}} \mathbf{y}^{(n)} = \mathbf{y}^{(n)} \label{eq:u}%\text{ for } n \in \{0, \ldots, N\} \notag 
\end{align} 
for $n \in \{0, \ldots, N\}$, which enters the next LN layer.  The mean across the elements in $\mathbf{u}^{(n)}$ is $\mu_2^{(n)}\approx \mu$ from \eqref{eq:meanY} and standard deviation is $\sigma_2^{(n)}\approx \sigma$ from \eqref{eq:stdY}. 
We set the weight parameters of this LN layer to $\sigma$ and bias parameters to $0$. Then, the LN output becomes,
\vspace{-0.1cm}
\begin{align} 
    \mathbf{p}^{(n)} &\triangleq \frac{\mathbf{u}^{(n)}-\mu_2^{(n)}}{\sigma_2^{(n)}} \odot \sigma \times  \mathbf{1}_D + 0 \times \mathbf{1}_D \approx \mathbf{y}^{(n)} \label{eq:p}
\end{align}
where $\mathbf{1}_D$ is a $D$-dimensional vector containing all $1$s. Next, MLP and LN layers within this encoder are designed to propogate embeddings from \eqref{eq:p} undistorted towards the next encoder (App. \ref{app:target_encoder}).

\vspace{-0.1cm}
\noindent
\textbf{Last Encoder (Encoder $S$). } 
Our goal in this stage is to ensure that the class token embedding contains the average of all token embeddings. As only the class token is used in the final layer  for loss prediction, this is critical for the final gradient to carry  information from all target tokens. 
Hence, the parameters within the MSA layer are designed to produce an attention matrix equal to  $\frac{1}{N+1}\mathbf{1}_{(N+1) \times (N+1)}$. 
This leads the MSA layer to output the average of all input token embeddings. Next, the parameters within the  MLP and LN layers are designed to propagate the average of the token embeddings added to the class token (through the residual connection). We denote this final embedding as $\mathbf{a}^{(0)} \approx \mathbf{p}^{(0)}+\frac{1}{N+1}\sum_{i=0}^{N}
\mathbf{p}^{(i)}$. 
The detailed designs are delegated to App. \ref{app:last_encoder}.
Then,
\vspace{-0.1cm}
\begin{align}
\frac{\partial \mathbf{a}^{(0)}_{d}}{\partial \mathbf{B}_{O,t}[:,n]} 
&\approx -\frac{1}{N+1} \frac{1}{\sigma^2 D} 
\sum_{i=0}^{N} \mathbf{y}^{(i)}_{d} \mathbf{y}^{(i)}_{2\mathcal{T}_t(n)+2} \mathbf{y}^{(i)} \notag \\
&\approx 
\begin{cases}
\frac{c_1^2}{(N+1)\sigma^2 D} \mathbf{y}^{(\mathcal{T}_t(n))}, & \text{if } d = 2\mathcal{T}_t(n)+1\\[2mm]
\mathbf{0}, & \text{otherwise}
\end{cases}
\label{eq:final}
\end{align}
 as demonstrated in App. \ref{App:2}.
%which will later allow us to recover the target token  $\mathbf{y}^{(\mathcal{T}_t(n))}$. 
Equation \eqref{eq:final} follows from the design of the position encoding vector in \eqref{eq:y} and \eqref{eq:Epos}, which ensures a large value for the multiplicative factor $-\mathbf{y}^{(i)}_{2\mathcal{T}_t(n)+1}\mathbf{y}^{(i)}_{2\mathcal{T}_t(n)+2}$ for the target token $i=\mathcal{T}_t(n)$. The detailed analysis is provided in App. \ref{App:2}.

\vspace{-0.1cm}
\noindent
{\bf Classification Layer. }
% {\bf Classification Layer. }
In the final layer, the class token embedding $\mathbf{a}^{(0)}$ is used  to produce the logits for each class. This is a linear layer with a weight matrix $\mathbf{W}_{CLS} \in \mathbb{R}^{D \times C}$, where $C$ is the number of classes. The bias parameters  are set to $0$.   
We set the weight matrix,
\vspace{-0.1cm}
\begin{align} \label{eq:weight_class}
\mathbf{W}_{CLS}[i,j] \triangleq \left\{ \begin{matrix} c_3 &\text{ if } i=2n'+1, n' \!\in\! \mathcal{T},  j=1\\
0 &\text{ otherwise } \end{matrix} \right.
\vspace{-0.3cm}
\end{align}
where  
$c_3$ is a large constant ($\sim 10^2$). 
Denote the output logit for class $i$ as $q_i$. Then, 
\vspace{-0.1cm}
\begin{align}
    q_1=\sum_{n' \in \mathcal{T}}\mathbf{a}^{(0)}_{2n'+1}\mathbf{W}_{CLS}[2n'+1,1] \label{eq:23}
\end{align} 
Let $z_i \triangleq e^{q_i}/\sum_{c=1}^Ce^{q_c}$ denote the predicted score for class $i$ after softmax. 
% After the softmax function, the predicted score for class $i$ is given  by $z_i \triangleq e^{q_i}/\sum_{c=1}^Ce^{q_c}$. 
The weights in \eqref{eq:weight_class} ensure that the score for class $1$ is $z_1\approx1$  and $0$ for all other classes, which allows the server to estimate the partial derivative of the loss function with respect to the logits.  
% The ground-truth label for class $i$ is $l_i=1$ if the fine-tuning sample belongs to class $i$ and $0$ otherwise. 
The loss function of user $u$ is given by 
    $\mathcal{L}_u=-\sum_{i=1}^Cl_i \log(z_i)$, where  $l_i=1$ if the fine-tuning sample belongs to class $i$ and $0$ otherwise. Then, 
% \begin{align}
%     \frac{\partial \mathcal{L}_u}{\partial q_1}&=-\frac{\partial}{\partial q_1}\sum_{i=1}^C l_i \log  z_i=z_1-l_1=1-l_1  \label{eq:24}
% \end{align}  
% belongs to class $i$ and $0$ otherwise.
the gradient with respect to the $n^{th}$ column of the  fine-tuning matrix $ \mathbf{B}_{O,t}$ in encoder $t$ can be written as, 
\vspace{-0.1cm}
% \begin{align}
%     \frac{\partial \mathcal{L}_u}{\partial \mathbf{B}_{O,t}[:,n]} = \sum_{n' \in \mathcal{T}}\frac{\partial \mathcal{L}_u}{\partial q_1}\frac{\partial q_1}{\partial \mathbf{a}^{(0)}_{2n'+1}}\frac{\partial \mathbf{a}^{(0)}_{2n'+1}}{\partial \mathbf{B}_{O,t}[:,n]}
%     %&=\!\!\sum_{n' \in \mathcal{T}}(z_1-l_1)\mathbf{W}_{CLS}[2n'+1,1]\frac{\partial\mathbf{a}^{(0)}_{2n'+1}}{\partial \mathbf{B}_{O,t}[:,n]} \notag\\
%     &\approx\!(z_1-l_1)c_3  \frac{c_1^2}{(N+1)\sigma^2 D} \mathbf{y}^{(\mathcal{T}_t(n))} \label{eq:chain_rule}
% \end{align} 
\begin{align}
\frac{\partial \mathcal{L}_u}{\partial \mathbf{B}_{O,t}[:,n]} 
&= \sum_{n' \in \mathcal{T}}
\frac{\partial \mathcal{L}_u}{\partial q_1} 
\frac{\partial q_1}{\partial \mathbf{a}^{(0)}_{2n'+1}}
\frac{\partial \mathbf{a}^{(0)}_{2n'+1}}{\partial \mathbf{B}_{O,t}[:,n]} \notag \\
&\approx (z_1 - l_1) c_3 
\frac{c_1^2}{(N+1)\sigma^2 D} 
\mathbf{y}^{(\mathcal{T}_t(n))}
\label{eq:chain_rule}
\end{align}
where \eqref{eq:chain_rule} follows from  \eqref{eq:final}, \eqref{eq:weight_class}, and \eqref{eq:23}. 
Note that the gradient in \eqref{eq:chain_rule}  is zero if the fine-tuning sample belongs to class $1$ (i.e.,  $l_1=1$),  and non-zero otherwise (when $l_1=0$).  
For the former, the server can predict the true class of the fine-tuning sample, and  set  the weights of a different class to $c_3$ in \eqref{eq:weight_class} in the next round, to obtain a non-zero gradient. Finally, the  
server can recover the target token embedding $\mathbf{y}^{(\mathcal{T}_t(n))}$  by dividing \eqref{eq:chain_rule} by a factor $c_3c_1^2 /((N+1)\sigma^2 D)$,  
and recover the target word embedding $\mathbf{x}^{(\mathcal{T}_t(n))}$ by subtracting the position encoding vector, $\mathbf{x}^{(\mathcal{T}_t(n))}=\mathbf{y}^{(\mathcal{T}_t(n))}- \mathbf{e}^{(\mathcal{T}_t(n))}$, 
% \begin{align}
% \mathbf{x}^{(\mathcal{T}_t(n))}=\mathbf{y}^{(\mathcal{T}_t(n))}- \mathbf{e}^{(\mathcal{T}_t(n))},   \notag 
% \end{align}
for $t\in[S-1], n\in\mathcal{T}_t$. 
Fig. \ref{fig:attack} illustrates token reconstruction from multiple encoders.

Note that our work adopts the standard adversarial server model in FL \cite{Fowl2023, sami2025gradient}, which attempts to recover user data by tampering with model parameters, without regard for preserving model accuracy. FL typically runs in the background without active monitoring, making malicious behavior less likely to be detected \cite{Ramaswamy2019, Fowl2023}. Since our attack is one-shot and reconstructs data in a single training round (e.g., at an early stage when accuracy is low), even if a user later notices degraded performance, the server may already have recovered the sensitive data. Investigating stealthier attacks that preserve training performance such as \cite{Feng2024} is an interesting future direction.

\begin{table*} %[h]
 \caption{Average BLEU and ROUGE-L scores over $100$ samples for different datasets and models.}
 \vspace{-0.2cm}
    \label{tab:100_samples}
 \footnotesize
% \scriptsize
    \centering
   % \vspace{-0.1cm}
    { %\scriptsize
    \fontsize{6pt}{7pt}\selectfont
    \begin{tabular}{lcccccccc}
        \toprule
        \textbf{Dataset} & \multicolumn{2}{c}{\textbf{RoBERTa-base}} &\multicolumn{2}{c}{\textbf{RoBERTa-large}} &\multicolumn{2}{c}{\textbf{BERT-base}} &\multicolumn{2}{c}{\textbf{BERT-large}}    \\
        \midrule
        & BLEU &ROUGE-L & BLEU &ROUGE-L & BLEU &ROUGE-L & BLEU &ROUGE-L\\
        \midrule
       AG's  & $1.0$ &$1.0$ & $1.0$ &$1.0$ & $1.0$ &$1.0$ &$0.82$ &$0.94$  \\
        %\cmidrule{2-3}
        DBPedia & $0.99$ &$1.0$ & $0.99$ &$1.0$  & $1.0$ &$1.0$ &$0.79$ &$0.92$ \\
        Yahoo  & $1.0$ &$1.0$ & $1.0$ &$1.0$ & $1.0$ &$1.0$ &$0.81$ &$0.94$\\
        Yelp  & $0.98$ &$1.0$ & $0.98$ &$1.0$ & $0.98$ &$1.0$ &$0.73$ &$0.91$\\
        TREC & $1.0$ &$1.0$ & $1.0$ &$1.0$ & $0.98$ &$1.0$&$0.82$ &$0.94$\\
        \bottomrule
    \end{tabular}
    }
\end{table*}
\vspace{0.2cm}
\begin{table*}[t]
 \caption{Recovered tokens from different datasets  (rank $r=4$, RoBERTa-base).  }
 \vspace{-0.2cm}
    \label{tab:comparison}
 \footnotesize
% \scriptsize
    \centering
    {
    %\scriptsize
    \fontsize{6pt}{7pt}\selectfont
    \begin{tabular}{lcc}
        \toprule
        \textbf{Dataset} & \multicolumn{2}{c}{\textbf{Fine-tuning sample}}    \\
        \midrule
        \multirow{3}{*}{AG's } & \multirow{1}{*}{Ground-truth } &Microsoft Sends Digital Business Cards New InterConnect 2004 software automatically updates contact info. \\
        % & &2004 software automatically updates contact info. \\
        \cmidrule{2-3}
        & \multirow{2}{*}{Recovered} &`Microsoft', `S', `ends', `Digital', `Business', `Cards', `New', `Inter',  \\
        & &`Connect', `2004', `software', `automatically', `updates', `contact', `info’,`.' \\
         \midrule
        \multirow{2}{*}{DBPedia} & Ground-truth  &Palacete de Belomonte is a historic palace in Porto Portugal. \\
        \cmidrule{2-3}
        & \multirow{1}{*}{Recovered} &`Pal', `ac', `ete', `de', `Bel', `omon', `te', `is', `a', `historic', `palace', `in', `Port', `o', `Portugal’, `.'\\
        %& &`in', `Port', `o', `Portugal’, `.' \\
        \midrule
        \multirow{3}{*}{Yahoo} & \multirow{1}{*}{Ground-truth} &Heavy water what is the role that heavy water plays in the nuclear explosion process? \\
        %& &in the nuclear explosion process ?\\
        \cmidrule{2-3}
        & \multirow{1}{*}{Recovered} &`Heavy', `water', `what', `is', `the', `role', `that', `heavy', `water', `plays', `in', `the', `nuclear', `explosion', `process', `?'  \\
       % & &`plays', `in', `the', `nuclear', `explosion', `process', `?'\\
        \midrule
        \multirow{2}{*}{Yelp} & \multirow{1}{*}{Ground-truth} &If you like eastern north carolina BBQ, you'll be in heaven.  \\
        \cmidrule{2-3}
        & \multirow{1}{*}{Recovered} &`If', `you', `like', `eastern', `north', `car', `olina', `BBQ', `,', `you', ` 'll ', `be', `in', `heaven', `.' \\
       % & &` 'll ', `be', `in', `heaven', `.'\\
        \midrule
        \multirow{3}{*}{TREC} & \multirow{1}{*}{Ground-truth} &What American League baseball team 's worst finish between 1926 and 1964 was fourth?\\
        %& &between 1926 and 1964 was fourth ? \\
        \cmidrule{2-3}
        & \multirow{2}{*}{Recovered} &`What', `American', `League', `baseball', `team', ` ' ', `s', `worst',\\
        & & `finish', `between', `1926', `and', `1964', `was', `fourth', `?' \\
        %& &`finish', `between', `1926', `and', `1964', `was', `fourth', `?'
%\\
        %\midrule
         %\midrule
        \bottomrule
    \end{tabular}
    }
    \vspace{-0.3cm}
\end{table*}

% We leverage fine-tuning gradients from $4$ encoders to recover $16$ tokens.

% Note that we can also use the gradients of the fine-tuning LoRA modules added to the value matrix, $\mathbf{W}_{V,t}$ for token reconstruction. For this, we can follow the same design principle in \eqref{eq:design_A}, \eqref{eq:design_B} for LoRA modules added to value matrix.
% %and set $\mathbf{W}_{V,t}^h$ to $\mathbf{0}$ for $h \in [H]$. 
% As we show in Section \ref{sec:exp}, from the gradients of LoRA modules added to both $\mathbf{W}_{V,t}$ and $\mathbf{W}_{O,t}$, tokens can be recovered. Exploring the attack by leveraging gradients of LoRA modules added to query, key matrices are left for future works.
\section{EXPERIMENTS} \label{sec:exp}

Our experiments seek to answer the following questions:
\begin{itemize}
\itemsep0em 
    \item How does \namespace perform in recovering tokens from text sequences?
    \item How does \namespace perform with different ranks of LoRA matrices?
    \item How can recovery rate be increased by leveraging  LoRA matrices from multiple encoders?
    \item How is the reconstruction performance affected by increasing the sequence length and batch size?
    \item How does \namespace perform in recovering images?
%\vspace{-0.2cm}
\end{itemize}

\vspace{-0.1cm}
\textbf{Setup.} 
We initially consider federated fine-tuning for text classification with $100$ users, each training locally and sending the LoRA gradients to the server. 
From each gradient, the server reconstructs
the fine-tuning data using \name.
Experiments are run on a 24 core AMD Ryzen with NVIDIA RTX4000.
\begin{figure}[t]
\centering
\subfigure[BLEU score]{%
  \label{fig:vary_r_Bleu}
  \includegraphics[width=0.48\linewidth]{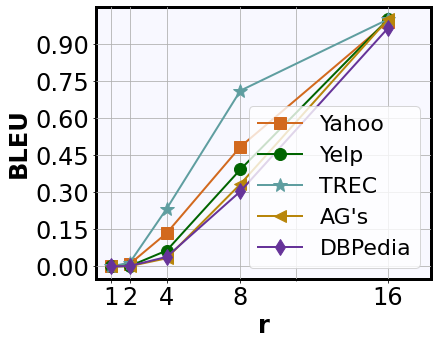}
}
\hspace{-0.4cm}
\subfigure[ROUGE-L score]{%
  \label{fig:var_r_Rouge}
  \includegraphics[width=0.48\linewidth]{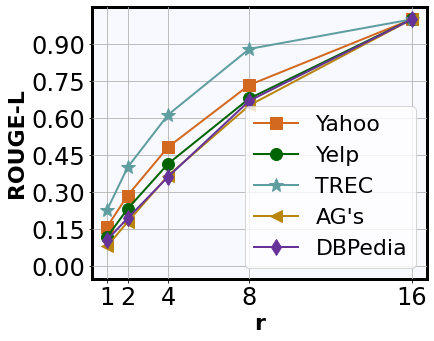}
}
\vspace{-0.4cm}
\caption{Reconstruction quality with varying $r$.}
\label{fig:vary_r}
\end{figure}
%\vspace{-0.1cm}

\textbf{Datasets and model architecture.} We demonstrate the results with multiple datasets, which includes Yahoo Answers Topics, Yelp Review, AG's News, DBPedia and TREC-6 \cite{Zhang2015, Li2002}. For the pretrained model, we consider BERT-base, BERT-large \cite{Jacob2019}, RoBERTa-base and RoBERTa-large \cite{Yinhan2019}.

\textbf{Performance metrics.} To evaluate the performance of \name,
%in terms of reconstruction quality, 
we measure BLEU and ROUGE-L scores \cite{Bleu2002, Rouge2004} between recovered and ground-truth texts. Higher score implies better reconstruction.

\textbf{Hyperparameters.} We use $c_1=10^2$, $c_2=3$ and $c_3=10^2$ in equations \eqref{eq:Epos}, \eqref{eq:weight_class}. Unless stated otherwise, we show the results with RoBERTa-base, rank $r=4$, and sequence length $16$.

\begin{figure}[ht]
\centering
\subfigure[BLEU score]{%
  \label{fig:vary_enc_Bleu}
  \includegraphics[width=0.48\linewidth]{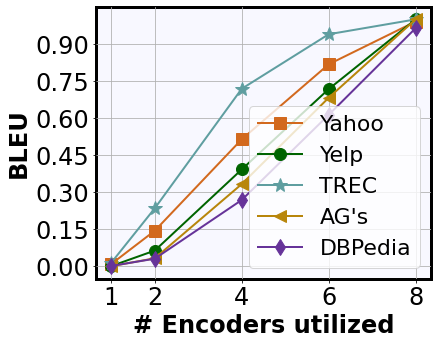}
}
\hspace{-0.4cm}
\subfigure[ROUGE-L score]{%
  \label{fig:var_enc_Rouge}
  \includegraphics[width=0.48\linewidth]{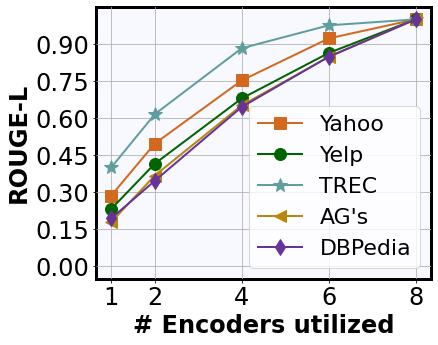}
}
\vspace{-0.4cm}
\caption{Utilizing multiple encoders ($r=2$).}
\label{fig:vary_enc}
\end{figure}
%vspace{0.2cm}
\textbf{Results.} 
In Table \ref{tab:100_samples}, we report the average scores across $100$ samples for different datasets and architectures under the same hyperparameters. As we observe, for all the models and datasets, \namespace achieves ROUGE-L and BLEU scores equal to (or close to) $1$, implying perfect (or near perfect) reconstruction. 
In Table \ref{tab:comparison}, we illustrate examples of tokens recovered from the shared gradients. 
From the gradients corresponding to each encoder, we recover $4$ tokens (as $r=4$). To recover all $16$ tokens, we use $4$ encoders.

\begin{figure*}[t] % Use figure* to span both columns
\centering
\subfigure[Ground-truth]{%
  \label{fig:image_GT}
  \includegraphics[width=0.45\textwidth]{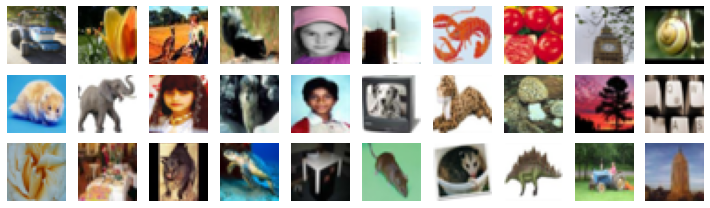}
}
\hspace{0.4cm}
\subfigure[Recovered]{%
  \label{fig:image_rec}
  \includegraphics[width=0.45\textwidth]{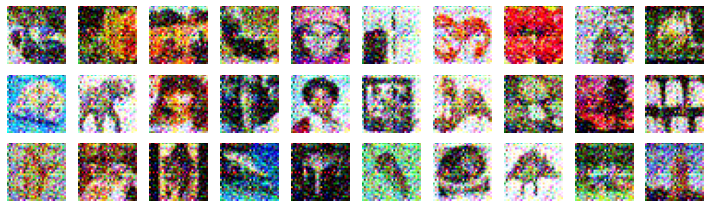}
}
\vspace{-0.3cm}
\caption{Image reconstruction ($r=4$) for CIFAR-100. }
\label{fig:image}
\vspace{-0.1cm}
\end{figure*}

\textbf{Reconstruction rate vs. rank of LoRA matrices.} In Fig. \ref{fig:vary_r}, we study the impact of rank $r$ on reconstruction for a single encoder. We observe that reconstruction quality enhances when $r$ increases.  This suggests adopting a small $r$ could be suffiicent to preserve privacy. However, in Fig. \ref{fig:vary_enc} we show that this is not the case. By deploying multiple  encoders, we can achieve the same BLEU and ROUGE-L scores for rank $r=2$ as for $r=16$ with a single encoder.

\begin{figure}[ht]
\centering
\subfigure[BLEU score]{%
  \label{fig:vary_seq_Bleu}
  \includegraphics[trim=9 8 8 6,clip,width=0.48\linewidth]{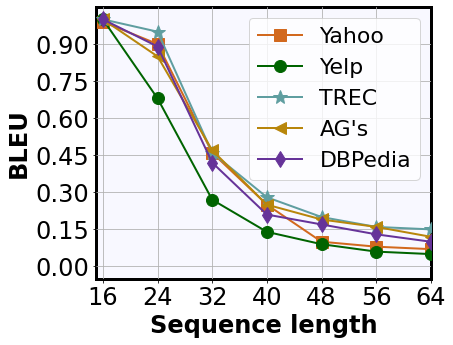}
}
\hspace{-0.4cm}
\subfigure[ROUGE-L score]{%
  \label{fig:vary_seq_Rouge}
  \includegraphics[trim=9 8 8 6,clip,width=0.48\linewidth]{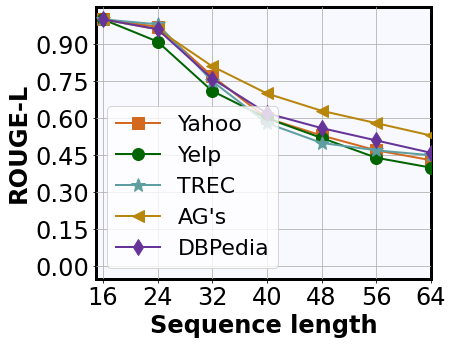}
}
\vspace{-0.4cm}
\caption{Reconstruction with varying sequence length.}
\label{fig:vary_seq}
\end{figure}

\begin{figure}[ht]
\centering
\subfigure[BLEU score]{%
  \label{fig:dager_bleu}
  \includegraphics[trim=9 8 8 6,clip,width=0.48\linewidth]{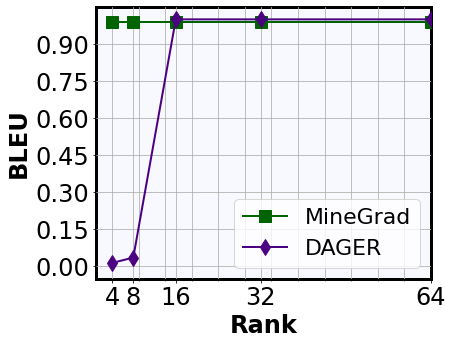}
}
\hspace{-0.4cm}
\subfigure[ROUGE-L score]{%
  \label{fig:dager_rouge}
  \includegraphics[trim=9 8 8 6,clip,width=0.48\linewidth]{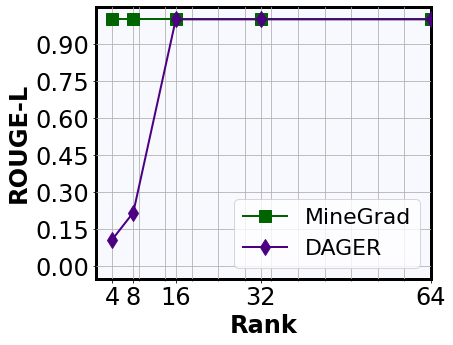}
}
\vspace{-0.4cm}
\caption{Reconstruction performance for DAGER vs. MineGrad (Yahoo Answers dataset).}
\label{fig:comparison}
\end{figure}

\textbf{Reconstruction rate vs. sequence length.} 
Fig. \ref{fig:vary_seq} shows our results for increasing sequence length with RoBERTa-base, using the same rank and number of encoders as Tables~\ref{tab:100_samples}–\ref{tab:comparison}. 
As expected, reconstruction performance degrades as the sequence length increases.
Performance can be further enhanced by tampering with pretrained word embeddings (App.~\ref{app:modified}).

\textbf{Comparison with DAGER \cite{Petrov2024}.} In Fig. \ref{fig:comparison}, we compare the performance of \namespace and DAGER.  
We observe that the reconstruction performance of DAGER heavily degrades when the rank is lower than the sequence length $16$. In contrast, \namespace retains the attack success even with a rank as small as $4$.

\textbf{Defense mechanisms.} In App. \ref{app:defense}, we present the attack performance against popular defenses %proposed to mitigate gradient inversion attacks, 
such as secure aggregation \cite{bonawitz2017practical}, pruning \cite{Yujun2018}, and noise \cite{Abadi2016}. 

\begin{table}[t]
\centering
\caption{Percentage of tokens recovered under different batch sizes across multiple datasets.}
\vspace{-0.15cm}
\footnotesize
\setlength{\tabcolsep}{6pt} % increase horizontal spacing
\renewcommand{\arraystretch}{0.9} % reduce vertical spacing
\begin{tabular}{@{}lcccc@{}}\toprule
\textbf{Batch size} & \textbf{8} & \textbf{16} & \textbf{32} & \textbf{64} \\
\midrule
Yahoo   & $99.7$ & $89.4$ & $64.9$ & $45.9$ \\
Yelp    & $99.5$ & $88.0$ & $65.3$ & $44.2$ \\
AGnews  & $99.8$ & $91.1$ & $66.3$ & $46.8$ \\
Trec    & $97.8$ & $85.8$ & $65.9$ & $52.2$ \\
DBpedia & $99.8$ & $89.9$ & $60.2$ & $41.6$ \\
\bottomrule
\end{tabular}
\label{tab:token_recovery}
\end{table}

\textbf{Reconstruction for a batch of sequences.} We next discuss how our attack performs for larger batch sizes (i.e., multiple sequences). Under this setup, we recover an average of multiple word embeddings from the gradient each associated with a particular sequence in the batch. From this average, the target word embeddings can be retrieved via a similarity-based search method across the vocabulary. Details are provided in App. \ref{App:4}. In Table \ref{tab:token_recovery}, we report the recovery performance. As we observe, even for a batch size of $64$, up to $52.2\%$ of the tokens can be recovered.

\begin{table}[ht]
\centering
\caption{LPIPS score across $100$ images.}
\label{table:LPIPS}
\vspace{-0.15cm}
\footnotesize
\setlength{\tabcolsep}{3pt}
\begin{tabular}{@{}lc c @{}}\toprule
& {\bf CIFAR-10} & {\bf CIFAR-100} \\
\cmidrule(lr){2-3}
LPIPS & $0.20 \pm 0.04$ & $0.21 \pm 0.06$ \\
\bottomrule
\end{tabular}
\end{table}

\textbf{Results for image recovery.} 
We next study how \namespace performs on vision transformers (ViT) to recover image samples \cite{Dosovitskiy2021}. 
In Table \ref{table:LPIPS}, we report the mean and standard deviation
for LPIPS scores \cite{Zhang2018} (between recovered and original images) across $100$ images from CIFAR-10 and CIFAR-100 datasets \cite{krizhevsky2009learning}.
In Fig. \ref{fig:image}, we provide the results for some sample images. We observe that the recovered images are %visually identifiable and 
 close to original images.

\textbf{Decoder-based architectures.}  In App.~~\ref{app:GPT2} we further provide our results with GPT-2 \cite{Radford2019}. 

\textbf{Ablation study.} 
In App. \ref{App:5}, we provide an ablation study by varying the hyperparameters $c_1$,  $c_2$ and additional examples of recovered images/text sequences in Apps.~\ref{app:img} and \ref{app:long_seq}.

App.~\ref{App:trainable} further demonstrates attack performance when LayerNorm and word embeddings are trainable. In App.~\ref{App:optimizers}, we consider the performance in the presence of adaptive optimizers, such as Adam and AdaGrad. 
Apps.~\ref{App:label_smoothing} and ~\ref{App:dropout}  demonstrate the performance under regularization, including Label Smoothing and Dropout.

\section{CONCLUSION} \label{sec:conc}
\vspace{-0.1cm}
We study how an adversary can recover sensitive fine-tuning data through malicious tampering with the pretrained model and fine-tuning LoRA modules.  
In addition to new privacy threats, we also demonstrate several new design principles that can be utilized in future studies. 
We show the efficacy of our attack across multiple models and datasets. Our results highlight the need for verifiable defense mechanisms for PEFT.

\section*{ACKNOWLEDGEMENT}
This work was supported in part by the NSF CAREER Award CCF-2144927, UCR OASIS Fellowship, and OUSD (R$\&$E)/RT$\&$L and was accomplished under Cooperative Agreement Number W911NF-20-2-0267. The views and conclusions contained in this document are those of the authors and should not be interpreted as representing the official policies, either expressed or implied, of the ONR, ARL and OUSD(R$\&$E)/RT$\&$L or the U.S. Government. The U.S. Government is authorized to reproduce and distribute reprints for Government purposes notwithstanding any copyright notation herein.   

%\newpage 

%\bibliographystyle{plainnat}
\bibliographystyle{apalike}
\bibliography{ref}

% %%%%%%%%%%%%%%%%%%%%%%%%%%%%%%%%%%%%%%%%%%%%%%%%%%%%%%%%%%%%%%%%%%%%%%%%%%%%%%%
% %%%%%%%%%%%%%%%%%%%%%%%%%%%%%%%%%%%%%%%%%%%%%%%%%%%%%%%%%%%%%%%%%%%%%%%%%%%%%%%
% APPENDIX
%%%%%%%%%%%%%%%%%%%%%%%%%%%%%%%%%%%%%%%%%%%%%%%%%%%%%%%%%%%%%%%%%%%%%%%%%%%%%%%
%%%%%%%%%%%%%%%%%%%%%%%%%%%%%%%%%%%%%%%%%%%%%%%%%%%%%%%%%%%%%%%%%%%%%%%%%%%%%%%
\clearpage
\appendix
\thispagestyle{empty}

\onecolumn

% \hspace{7cm}
{\bf \Large APPENDIX} 
\section{BROADER IMPACT} \label{App:impact}
We demonstrate how fine-tuning samples can be recovered from the LoRA gradients in an FL setup. In practice, an adversarial server can get access to private fine-tuning data of a victim user through malicious tampering with the fine-tuning protocol. Pretrained models can be downloaded from a compromised source without a stringent verification protocol. In FL, users implicitly trust the models shared by the server. The server can leverage this scenario and send a poisonous pretrained model to the users. In addition, users rely on the server for receiving the updated fine-tuning parameters in each training round. By deploying a poisonous design of the pretrained model and fine-tuning modules, the server can successfully captivate the fine-tuning data inside a user's local gradient. Without adopting further defense mechanisms, such fine-tuning does not ensure privacy. Therefore, it is important to further adopt formal  mechanisms to guarantee authenticity of the models received from the server. By demonstrating the recoverability of sensitive fine-tuning samples with lightweight analytical attacks, without access to heavy computational resources, we hope to motivate new incentives towards employing robust defense mechanisms with provable guarantees in practice.
\section{DETAILS OF THE ATTACK DESIGN}\label{App:1}
In this section, we describe the malicious design details of the 
encoders. 

\subsection{Design of Target Encoders} % $t \in [S-1]$}
\label{app:target_encoder}

{\bf MSA layer.} In the MSA layer, we have the embeddings $\mathbf{y}^{(n)}$ as inputs. 
For head $h \in [H]$,  we denote the query, key and value weight matrices of a target encoder $t \in [S-1]$ as $\mathbf{W}_{Q,t}^h, \mathbf{W}_{K,t}^h$ and $\mathbf{W}_{V,t}^h$, and biases  as $\mathbf{b}_{Q,t}^h, \mathbf{b}_{K,t}^h$ and $\mathbf{b}_{V,t}^h$, respectively, which are designed as,
\vspace{-0.05cm}\begin{align} \mathbf{W}_{Q,t}^h&=\mathbf{W}_{K,t}^h=\mathbf{W}_{V,t}^h= \mathbf{I}_{\overline{D} \times \overline{D}}  \label{eq:weight_MSA}\\
    &\mathbf{b}_{Q,t}^h= \mathbf{b}_{K,t}^h= \mathbf{b}_{V,t}^h=\mathbf{0} \label{eq:bias_MSA}  
\end{align} 
The LoRA matrices added to query, key and value weights are set to $\mathbf{0}$. Then, query, key and value for head $h$ are computed as,
\vspace{-0.05cm}\begin{align}
    \mathbf{Q}_t^h &\triangleq \begin{bmatrix}\mathbf{W}_{Q,t}^h (\mathbf{y}^{(0)})_h &\cdots &\mathbf{W}_{Q,t}^h (\mathbf{y}^{(N)})_h \end{bmatrix}^{\text{T}}\notag\\
    \mathbf{K}_t^h &\triangleq \begin{bmatrix}\mathbf{W}_{K,t}^h (\mathbf{y}^{(0)})_h &\cdots &\mathbf{W}_{K,t}^h (\mathbf{y}^{(N)})_h \end{bmatrix}^{\text{T}}\notag\\ 
    \mathbf{V}^h &\triangleq \begin{bmatrix}\mathbf{W}_{V,t}^h (\mathbf{y}^{(0)})_h &\cdots &\mathbf{W}_{V,t}^h (\mathbf{y}^{(N)})_h \end{bmatrix} ^{\text{T}} 
\end{align} 
Now, following \eqref{eq:condn1}, the self-attention output for head $h$ becomes,
\begin{align}
    SA_h([\mathbf{y}^{(0)} \hspace{0.2cm}\cdots \hspace{0.2cm}\mathbf{y}^{(N)}])  \triangleq softmax(\mathbf{Q}_t^h(\mathbf{K}_t^h)^{\text{T}}/\sqrt{\overline{D}}) \mathbf{V}_t^h
     &\cong \mathbf{I}_{(N+1) \times (N+1)} \mathbf{V}_t^h \notag\\
     &=\begin{bmatrix}
         (\mathbf{y}^{(0)})_h & 
        \cdots&
        (\mathbf{y}^{(N)})_h
    \end{bmatrix}^{\text{T}}
\end{align}
Hence, our malicious design of query, key, value weights, biases and LoRA matrices along with the condition from  \eqref{eq:condn1} produces an identity mapping of the input embeddings. Finally, as in \eqref{eq:MSAop}, we have,
\begin{align}
    MSA([\mathbf{y}^{(0)} \hspace{0.2cm}\cdots \hspace{0.2cm} \mathbf{y}^{(N)}]) 
   &=\hspace{-0.1cm}=\begin{bmatrix}
         (\mathbf{y}^{(0)}) & 
        \cdots&
        (\mathbf{y}^{(N)})
    \end{bmatrix}^{\text{T}}\mathbf{B}_{O,t}\mathbf{A}_{O,t} 
\end{align} 

\textbf{MLP layer.} Embeddings $\mathbf{p}^{(n)}$ from \eqref{eq:p} enters the MLP layer.  
We set all the pretrained weight parameters in the MLP layer to $0$ to produce zero output. Then, through the residual connection, input to the next LN layer is equal to $\mathbf{p}^{(n)}$ from \eqref{eq:p}. By following the same design as LN layer parameters in \eqref{eq:p}, the output of this LN layer is approximately equal to $\mathbf{p}^{(n)}$. 

\subsection{Design of the Last Encoder} % (Encoder $S$)} 
\label{app:last_encoder}

{\bf MSA layer.} 
%We now discuss how we obtain \eqref{eq:SA_output_2}. 
For the MSA layer in the last encoder (Encoder $S$), we have embeddings $\mathbf{p}^{(n)}$ as inputs for $n \in \{0, \ldots, N\}$. For head $h \in [H]$,  the query, key and value parameters $\mathbf{W}_{Q,S}^h, \mathbf{W}_{K,S}^h$ and $\mathbf{W}_{V,S}^h$ 
are designed as,
\vspace{-0.05cm}\begin{align} \mathbf{W}_{Q,S}^h&=\mathbf{W}_{K,S}^h= \mathbf{0}_{\overline{D} \times \overline{D}}, \hspace{0.2cm} \mathbf{W}_{V,S}^h=\mathbf{I}_{\overline{D} \times \overline{D}}  \end{align}
and biases $\mathbf{b}_{Q,S}^h, \mathbf{b}_{K,S}^h$, $\mathbf{b}_{V,S}^h$ are designed as,
\begin{align}
    &\mathbf{b}_{Q,S}^h= \mathbf{b}_{Q,S}^h= \mathbf{b}_{V,S}^h=\mathbf{0} %%\label{eq:bias_MSA}  
\end{align}  
The LoRA modules added to query, key and value weights are set to $\mathbf{0}$.
Then, query, key and values for head $h$ are,
\vspace{-0.05cm}\begin{align}
    \mathbf{Q}_S^h &\triangleq \begin{bmatrix}\mathbf{W}_{Q,S}^h (\mathbf{p}^{(0)})_h &\cdots &\mathbf{W}_{Q,S}^h (\mathbf{p}^{(N)})_h \end{bmatrix}^{\text{T}}= \mathbf{0}\\
    \mathbf{K}_S^h &\triangleq \begin{bmatrix}\mathbf{W}_{K,S}^h (\mathbf{p}^{(0)})_h &\cdots &\mathbf{W}_{K,S}^h (\mathbf{p}^{(N)})_h \end{bmatrix}^{\text{T}}= \mathbf{0}\\ 
    \mathbf{V}_S^h &\triangleq \begin{bmatrix}\mathbf{W}_{V,S}^h (\mathbf{p}^{(0)})_h &\cdots &\mathbf{W}_{V,S}^h (\mathbf{p}^{(N)})_h \end{bmatrix}^{\text{T}} = \begin{bmatrix} (\mathbf{p}^{(0)})_h &\cdots &(\mathbf{p}^{(N)})_h \end{bmatrix}^{\text{T}} 
\end{align} 
Hence, the attention matrix becomes a matrix with all elements equal to $\frac{1}{N+1}$,
\begin{equation}
  softmax(\mathbf{Q}_S^h(\mathbf{K}_S^h)^{\text{T}}/\sqrt{\overline{D}}) 
    \cong \frac{1}{N+1}\mathbf{1}_{(N+1) \times (N+1)} \notag
\end{equation}
where $\mathbf{1}_{(N+1) \times (N+1)}$ represents a matrix containing all $1$s.
The self-attention output for head $h$ is computed as,
\begin{align}
SA_h([\mathbf{p}^{(0)} \hspace{0.2cm} \cdots \hspace{0.2cm} \mathbf{p}^{(N)}]) 
&\triangleq softmax(\mathbf{Q}_S^h(\mathbf{K}_S^h)^{\text{T}}/\sqrt{\overline{D}})\mathbf{V}_S^h \notag\\
&=
\hspace{-0.05cm}\begin{bmatrix} \frac{1}{N+1}\hspace{-0.05cm}\sum_{i=0}^{N}(\mathbf{p}^{(i)})_h \hspace{0.15cm}\cdots \hspace{0.15cm}  \frac{1}{N+1}\hspace{-0.05cm}\sum_{i=0}^{N}(\mathbf{p}^{(i)})_h\end{bmatrix}^{\text{T}}
\end{align}

We then set $\mathbf{W}_{O,S} = \mathbf{I}_{D \times D}$, and LoRA modules $\mathbf{A}_{O,S}, \mathbf{B}_{O,S}$ to $\mathbf{0}$, after which  the MSA output becomes,
\begin{align}
    &MSA([\mathbf{p}^{(0)} \hspace{0.2cm} \cdots \hspace{0.2cm} \mathbf{p}^{(N)}])\notag\\
    &= \begin{bmatrix} \frac{1}{N+1}\sum_{i=0}^{N}\mathbf{p}^{(i)} \hspace{0.2cm}\cdots \hspace{0.2cm} \frac{1}{N+1}\sum_{i=0}^{N}\mathbf{p}^{(i)}\end{bmatrix}^{\text{T}}
\end{align} 
and  ensures the average embeddings propagate undistorted. 
After the residual connection, class token embedding is,
\begin{align} \label{eq:v}
    \mathbf{v}^{(0)} \triangleq \mathbf{p}^{(0)}+\frac{1}{N+1}\sum_{i=0}^{N}
\mathbf{p}^{(i)}
\end{align}

 We design the parameters of the subsequent LN layer following the same intuition as the LN layer from the preceding encoders, which produces an identity mapping. 

\textbf{MLP layer.}  All weight parameters in the MLP layer are set to $0$ to produce zero output. After going through the residual connection and the LN layer (similar design intuition as the previous LN layer), the output of the last encoder is,  
\begin{align} \label{eq:a}
    \mathbf{a}^{(n)} 
 \approx \mathbf{v}^{(n)} \text{ for } n\in\{0, \ldots, N\}
 \end{align} 

\section{GRADIENT COMPUTATION}\label{App:2}
In this section, we describe how we obtain \eqref{eq:final}. Since $\mathbf{a}^{(0)} \approx \mathbf{v}^{(0)}$ from \eqref{eq:a}, we have for $n' \in \mathcal{T}$ (the set of target tokens),
\begin{align}
    \frac{\partial\mathbf{a}^{(0)}_{2n'+1}}{\partial \mathbf{B}_{O,t}[:,n]} &\approx \frac{\partial \mathbf{v}^{(0)}_{2n'+1}}{ \partial \mathbf{B}_{O,t}[:,n]}\\
    &= \frac{\partial (\mathbf{p}^{(0)}_{2n'+1}+\frac{1}{N+1}\sum_{i=0}^{N}\mathbf{p}^{(i)}_{2n'+1})}{ \partial \mathbf{B}_{O,t}[:,n]} \label{eq:v=p}\\
    &=\frac{\partial \mathbf{p}_{2n'+1}^{(0)}}{\partial \mathbf{B}_{O,t}[:,n]}+\frac{\partial \frac{1}{N+1}\sum_{i=0}^{N}\mathbf{p}^{(i)}_{2n'+1}}{\partial \mathbf{B}_{O,t}[:,n]} \label{eq:der_a}
    %&\triangleq t_1 + t_2 +t_3
\end{align}
where \eqref{eq:v=p} follows from \eqref{eq:v}. From \eqref{eq:p}, $\mathbf{p}_{2n'+1}^{(i)}=\frac{\mathbf{y}_{2n'+1}^{(i)}+\mathbf{A}_{O,t}^{\text{T}}[2n'+1,:]\mathbf{B}_{O,t}^{\text{T}} \mathbf{y}^{(i)}-\mu_2^{(i)}}{\sigma_2^{(i)}}\sigma$. Hence,
\begin{align} \label{eq:terms_in_p}
    \frac{\partial \mathbf{p}_{2n'+1}^{(i)}}{\partial \mathbf{B}_{O,t}[:,n]} &= \frac{\partial \frac{\mathbf{y}_{2n'+1}^{(i)}+\mathbf{A}_{O,t}^{\text{T}}[2n'+1,:]\mathbf{B}_{O,t}^{\text{T}} \mathbf{y}^{(i)}-\mu_2^{(i)}}{\sigma_2^{(i)}}\sigma}{\partial \mathbf{B}_{O,t}[:,n]} 
\end{align}
Note that the mean of $\mathbf{u}^{(i)}$ in \eqref{eq:u} is,
\begin{align} \label{eq:mean2=0}
   \mu_2^{(i)} = \frac{1}{D}\sum_{d=1}^D(\mathbf{y}_{d}^{(i)}+\mathbf{A}_{O,t}^{\text{T}}[d,:]\mathbf{B}_{O,t}^\text{T}\mathbf{y}^{(i)}) =\frac{1}{D}\sum_{d=1}^D\mathbf{y}_{d}^{(i)} \approx 0
\end{align}
which follows from \eqref{eq:meanY}. Now, 
\begin{align}
    \frac{\partial \mu_2^{(i)}}{\partial \mathbf{B}_{O,t}[:,n]}
  %\label{eq:mean=0}\\
    &=\frac{\partial }{\partial \mathbf{B}_{O,t}[:,n]} \frac{1}{D}\sum_{d=1}^D(\mathbf{y}_{d}^{(i)}+\mathbf{A}_{O,t}^{\text{T}}[d,:]\mathbf{B}_{O,t}^\text{T}\mathbf{y}^{(i)})\\
    &=\frac{\partial }{\partial \mathbf{B}_{O,t}[:,n]} \frac{1}{D}\sum_{d=1}^D\mathbf{A}_{O,t}^{\text{T}}[d,:]\mathbf{B}_{O,t}^\text{T}\mathbf{y}^{(i)}= \frac{1}{D}\mathbf{y}^{(i)} \label{eq:use_derATBT}
\end{align}
where \eqref{eq:use_derATBT} follows from \eqref{eq:der_ATBT}. The standard deviation of $\mathbf{u}^{(i)}$ in \eqref{eq:u} is,
\begin{align}
    \sigma_2^{(i)} &\triangleq \sqrt{\frac{1}{D}\sum_{d=1}^D(\mathbf{y}_d^{(i)}+\mathbf{A}_{O,t}^\text{T}[d,:]\mathbf{B}_{O,t}^\text{T}\mathbf{y}^{(i)}-\mu_2^{(i)})^2} \approx \sqrt{\frac{1}{D}\sum_{d=1}^D(\mathbf{y}_d^{(i)}-0)^2} \approx \sigma \label{eq:approx3}
\end{align}
which follows from \eqref{eq:stdY}. Next,
 \begin{align}
    \frac{\partial \frac{\mu_2^{(i)}}{\sigma_2^{(i)}}}{\partial \mathbf{B}_{O,t}[:,n]}&=\frac{(\sigma_2^{(i)}) \frac{\partial \mu_2^{(i)}}{\partial \mathbf{B}_{O,t}[:,n]}-\mu_2^{(i)}\frac{\partial \sigma_2^{(i)}}{\partial \mathbf{B}_{O,t}[:,n]}}{(\sigma_2^{(i)})^2}\\
    &\approx \frac{1}{\sigma_2^{(i)}}\frac{\partial \mu_2^{(i)}}{\partial \mathbf{B}_{O,t}[:,n]} \label{eq:mean=0}\\
    &=\frac{1}{\sigma_2^{(i)}}\frac{\partial }{\partial \mathbf{B}_{O,t}[:,n]} \frac{1}{D}\sum_{d=1}^D(\mathbf{y}_d^{(i)}+\mathbf{A}_{O,t}^{\text{T}}[d,:]\mathbf{B}_{O,t}^\text{T}\mathbf{y}^{(i)})\\
    &=\frac{1}{\sigma_2^{(i)}}\frac{\partial }{\partial \mathbf{B}_{O,t}[:,n]} \frac{1}{D}\sum_{d=1}^D\mathbf{A}_{O,t}^{\text{T}}[d,:]\mathbf{B}_{O,t}^\text{T}\mathbf{y}^{(i)}\\
    &\approx \frac{1}{\sigma}\frac{1}{D}\mathbf{y}^{(i)} \label{eq:der_mu}
\end{align}
where \eqref{eq:mean=0} follows since $\mu_2^{(i)}\approx 0$ in \eqref{eq:mean2=0} and \eqref{eq:der_mu} follows from \eqref{eq:der_ATBT} and \eqref{eq:approx3}. 
%Since $\sigma_2^{(i)}$ is a function of $\mathbf{B}_{O,t}$, we consider its derivative with respect to $\mathbf{B}_{O,t}[:,n]$.
Next,
\begin{align}
&\frac{\partial \frac{\mathbf{y}_{2n'+1}^{(i)}}{\sigma_2^{(i)}}}{\partial \mathbf{B}_{O,t}[:,n]}\notag\\
&= \mathbf{y}_{2n'+1}^{(i)}\frac{\partial}{\partial \mathbf{B}_{O,t}[:,n]}\frac{1}{\sigma_2^{(i)}}\\
   &=-\mathbf{y}_{2n'+1}^{(i)}\frac{1}{(\sigma_2^{(i)})^2}\frac{\partial \sigma_2^{(i)}}{\partial  \mathbf{B}_{O,t}[:,n]}\\
   &=-\mathbf{y}_{2n'+1}^{(i)}\frac{1}{(\sigma_2^{(i)})^2}\frac{\partial ((\sigma_2^{(i)})^2)^{\frac{1}{2}}}{\partial  \mathbf{B}_{O,t}[:,n]}\\
   &=-\frac{1}{2}\mathbf{y}_{2n'+1}^{(i)}\frac{1}{(\sigma_2^{(i)})^3}\frac{\partial ((\sigma_2^{(i)})^2)}{\partial  \mathbf{B}_{O,t}[:,n]}\\
   &=-\frac{1}{2}\mathbf{y}_{2n'+1}^{(i)}\frac{1}{(\sigma_2^{(i)})^3}\frac{\partial}{\partial  \mathbf{B}_{O,t}[:,n]}\frac{1}{D}\sum_{d=1}^D(\mathbf{y}_{d}^{(i)}+\mathbf{A}_{O,t}^{\text{T}}[d,:]\mathbf{B}_{O,t}^\text{T}\mathbf{y}^{(i)}-\mu_2^{(i)})^2\\
   &=-\frac{1}{2}\mathbf{y}_{2n'+1}^{(i)}\frac{1}{(\sigma_2^{(i)})^3}\frac{1}{D}2\sum_{d=1}^D(\mathbf{y}_{d}^{(i)}+\mathbf{A}_{O,t}^{\text{T}}[d,:]\mathbf{B}_{O,t}^\text{T}\mathbf{y}^{(i)}-\mu_2^{(i)}) \notag\\
   &\hspace{5cm} \times \frac{\partial}{\partial  \mathbf{B}_{O,t}[:,n]}(\mathbf{y}_{d}^{(i)}+\mathbf{A}_{O,t}^{\text{T}}[d,:]\mathbf{B}_{O,t}^\text{T}\mathbf{y}^{(i)}-\mu_2^{(i)})\\
   &\approx-\mathbf{y}_{2n'+1}^{(i)}\frac{1}{(\sigma_2^{(i)})^3}\frac{1}{D}\sum_{d=1}^D\mathbf{y}_{d}^{(i)}\frac{\partial}{\partial  \mathbf{B}_{O,t}[:,n]}(\mathbf{A}_{O,t}^{\text{T}}[d,:]\mathbf{B}_{O,t}^\text{T}\mathbf{y}^{(i)}-\mu_2^{(i)}) \label{eq:approx1}\\
   &=\mathbf{y}_{2n'+1}^{(i)}\frac{1}{(\sigma_2^{(i)})^3D^2}\sum_{d=1, d\neq 2\mathcal{T}_t(n)+2}^D\hspace{-0.5cm}\mathbf{y}_{d}^{(i)}\begin{bmatrix}\mathbf{y}^{(i)}_1\\
    \mathbf{y}^{(i)}_2\\
    \vdots\\
    \mathbf{y}^{(i)}_D
    \end{bmatrix}- \frac{\mathbf{y}_{2n'+1}^{(i)}}{(\sigma_2^{(i)})^3D}\mathbf{y}_{2\mathcal{T}_t(n)+2}^{(i)}\begin{bmatrix}\mathbf{y}^{(i)}_1\\
    \mathbf{y}^{(i)}_2\\
    \vdots\\
    \mathbf{y}^{(i)}_D
    \end{bmatrix} \notag\\
    &\hspace{6cm}+\mathbf{y}_{2n'+1}^{(i)}\frac{1}{(\sigma_2^{(i)})^3}\frac{1}{D^2}\mathbf{y}_{2\mathcal{T}_t(n)+2}^{(i)}\begin{bmatrix}\mathbf{y}^{(i)}_1\\
    \mathbf{y}^{(i)}_2\\
    \vdots\\
    \mathbf{y}^{(i)}_D
    \end{bmatrix} \label{eq:3terms}\\
    & \approx \mathbf{y}_{2n'+1}^{(i)}\frac{1}{\sigma^3D^2}\sum_{d=1, d\neq 2\mathcal{T}_t(n)+2}^D\hspace{-0.5cm}\mathbf{y}_{d}^{(i)}\begin{bmatrix}\mathbf{y}^{(i)}_1\\
    \mathbf{y}^{(i)}_2\\
    \vdots\\
    \mathbf{y}^{(i)}_D
    \end{bmatrix}- \frac{\mathbf{y}_{2n'+1}^{(i)}}{\sigma^3D}\mathbf{y}_{2\mathcal{T}_t(n)+2}^{(i)}\begin{bmatrix}\mathbf{y}^{(i)}_1\\
    \mathbf{y}^{(i)}_2\\
    \vdots\\
    \mathbf{y}^{(i)}_D
    \end{bmatrix} \notag\\
    & \hspace{6cm}+\mathbf{y}_{2n'+1}^{(i)}\frac{1}{\sigma^3}\frac{1}{D^2}\mathbf{y}_{2\mathcal{T}_t(n)+2}^{(i)}\begin{bmatrix}\mathbf{y}^{(i)}_1\\
   \mathbf{y}^{(i)}_2\\
    \vdots\\
    \mathbf{y}^{(i)}_D
    \end{bmatrix} \label{eq:approx_A}
\end{align}
where \eqref{eq:approx1} follows since $\mu_2^{(i)}\approx 0$ and $\mathbf{A}_{O,t}^{\text{T}}[d,:]\mathbf{B}_{O,t}^\text{T}=0$ for all $d \in [D]$. Equation \eqref{eq:3terms} follows from \eqref{eq:der_ATBT} and \eqref{eq:use_derATBT}. Equation \eqref{eq:approx_A} follows since $\sigma_2^{(i)} \approx \sigma$ in \eqref{eq:approx3}. In \eqref{eq:approx_A}, compared to second term, the first and third terms have negligible impact since they are multiplied by an additional factor of $\frac{1}{D}$, which is very small for $D=768$ (base architecture) and $D=1024$ (large architecture).
For the second term, if $i=\mathcal{T}_t(n)$ and $n'=\mathcal{T}_t(n)$,
then
\begin{align}
    - \mathbf{y}_{2n'+1}^{(i)}\frac{1}{\sigma^3}\frac{1}{D}\mathbf{y}_{2\mathcal{T}_t(n)+2}^{(i)}\begin{bmatrix}\mathbf{y}^{(i)}_1\\
    \mathbf{y}^{(i)}_2\\
    \vdots\\
    \mathbf{y}^{(i)}_D
    \end{bmatrix} \approx c_1^2\frac{1}{\sigma^3}\frac{1}{D}
    \mathbf{y}^{(i)}
\end{align}
which follows from the design of position encoding vectors in \eqref{eq:Epos}.
For all other values of $i$ and $n'$, the term is negligible. 
Thus,
\begin{align}
    \frac{\partial \frac{\mathbf{y}_{2n'+1}^{(i)}}{\sigma_2^{(i)}}}{\partial \mathbf{B}_{O,t}[:,n]}&\approx \left\{ 
    \begin{matrix}c_1^2\frac{1}{\sigma^3}\frac{1}{D}\mathbf{y}^{(i)} &\text{ if } i=\mathcal{T}_t(n), n'=\mathcal{T}_t(n)\\
    \text{ negligible } &\text{ otherwise }
    \end{matrix} \right. \label{eq:der_y_divide_sigma}
\end{align}

Next, as we know from \eqref{eq:AB=0}, $\mathbf{A}_{O,t}^{\text{T}}[d,:]\mathbf{B}_{O,t}^{\text{T}} \mathbf{y}^{(i)}=0$ for any $d \in [D]$, therefore,
\begin{align}
    \frac{\partial \frac{\mathbf{A}_{O,t}^{\text{T}}[d,:]\mathbf{B}_{O,t}^{\text{T}} \mathbf{y}^{(i)}}{\sigma_2^{(i)}}}{\partial \mathbf{B}_{O,t}[:,n]}= \frac{1}{\sigma_2^{(i)}}\frac{\partial \mathbf{A}_{O,t}^{\text{T}}[d,:]\mathbf{B}_{O,t}^{\text{T}} \mathbf{y}^{(i)}}{\partial \mathbf{B}_{O,t}[:,n]}&=\left\{ \begin{matrix}\frac{\mathbf{y}^{(i)}}{\sigma_2^{(i)}} \hspace{1cm} &\text{ if } d = 2\mathcal{T}_t(n)+2\\
    \mathbf{0}_D \hspace{1cm} &\text{ otherwise }
    \end{matrix} \right. \\
    &\approx \left\{ \begin{matrix}\frac{\mathbf{y}^{(i)}}{\sigma} \hspace{1cm} &\text{ if } d=2\mathcal{T}_t(n)+2\\
    \mathbf{0}_D \hspace{1cm} &\text{ otherwise }
    \end{matrix} \right. \label{eq:der_ATBTsigma}
\end{align}
Hence, in \eqref{eq:terms_in_p}, 
\begin{align}
    \frac{\partial \frac{\mathbf{A}_{O,t}^{\text{T}}[2n'+1,:]\mathbf{B}_{O,t}^{\text{T}} \mathbf{y}^{(i)}}{\sigma_2^{(i)}}}{\partial \mathbf{B}_{O,t}[:,n]}= \mathbf{0}
\end{align}

Now, from  \eqref{eq:use_derATBT} and \eqref{eq:der_y_divide_sigma}, for $i=\mathcal{T}_t(n)$ and $n'=\mathcal{T}_t(n)$,
$\frac{\partial \frac{\mathbf{y}_{2n'+1}^{(i)}}{\sigma_2^{(i)}}}{\partial \mathbf{B}_{O,t}[:,n]}>>\frac{\partial \frac{\mu_2^{(i)}}{\sigma_2^{(i)}}}{\partial \mathbf{B}_{O,t}[:,n]}$ since $\frac{c_1^2}{\sigma^3D}>>\frac{1}{\sigma D}$ (as observed from our experiments). Therefore, from \eqref{eq:der_y_divide_sigma} and \eqref{eq:der_ATBTsigma}; \eqref{eq:terms_in_p} can be rewritten as,
\begin{align}
    \frac{\partial \mathbf{p}_{2n'+1}^{(i)}}{\partial \mathbf{B}_{O,t}[:,n]}&\approx \left\{ 
    \begin{matrix}c_1^2\frac{1}{\sigma^2}\frac{1}{D}\mathbf{y}^{(i)} &\text{ if } i=\mathcal{T}_t(n), n'=\mathcal{T}_t(n)\\
    \text{ negligible } &\text{ otherwise }
    \end{matrix} \right. 
\end{align}
Finally, in \eqref{eq:der_a},
\begin{align}
    \frac{\partial \mathbf{a}_{2n'+1}^{(0)}}{\partial \mathbf{B}_{O,t}[:,n]}&\approx \left\{ 
    \begin{matrix}\frac{1}{N+1}c_1^2\frac{1}{\sigma^2}\frac{1}{D}\mathbf{y}^{(\mathcal{T}_t(n))} &\text{ if } i=\mathcal{T}_t(n), n'=\mathcal{T}_t(n)\\
    \text{ negligible } &\text{ otherwise }
    \end{matrix} \right. 
\end{align}
from which the target token $\mathbf{y}^{(\mathcal{T}_t(n))}$ can be retrieved. Similarly, different tokens can be recovered by leveraging different columns $n \in [r]$ of matrix $\mathbf{B}_{O,t}$ and different encoders $t \in [S-1]$, where $r$ is the rank of LoRA matrices and $S$ is the total number of encoders in the architecture.

\begin{table*}[t]
  \caption{Recovered texts of the two target users from the gradient aggregate with respect to LoRA matrices. The recovered words are highlighted.}
  %The mispredicted words are colored in red. }
    \label{tab:secagg}
    \centering
     \fontsize{8pt}{9pt}\selectfont
    \begin{tabular}{lcc|c}
        \toprule
        \textbf{Sequence length}   &\textbf{User}  & \multicolumn{2}{c}{\textbf{Fine-tuning sample}}    \\
         \midrule
       \multirow{8}{*}{ 16} &  \multirow{4}{*}{ 1} & \multirow{2}{*} {Ground-truth}  &Will the raise in the minimum wage help you \\
       & & &or anyone you know over 19 years   \\
        \cmidrule{3-4}
        &   & \multirow{2}{*}{Recovered}  &\colorbox{lime}{Will the raise in the minimum wage help you} \\
        & & & \colorbox{lime}{or anyone you know over 19 years}   \\
         \cmidrule{2-4}
        & \multirow{4}{*}{ 2} & \multirow{2}{*} {Ground-truth}  &iv been doing alot more excersise and \\
        & & & eating approx 1200 cals per day\\
        \cmidrule{3-4}
      &  & \multirow{2}{*}{Recovered} &\colorbox{lime}{iv been doing alot more excersise and} \\
      & & & \colorbox{lime}{eating approx 1200 cals per day}
 \\
\midrule
        \multirow{8}{*}{ 32} &  \multirow{4}{*}{ 1} & \multirow{2}{*} {Ground-truth}  &Will the raise in the minimum wage help you \\
        & & & or anyone you know over 19 years   \\
        %\cmidrule{2-3}
         \cmidrule{3-4}
        &  & \multirow{2}{*}{Recovered} &\colorbox{lime}{Will}the\colorbox{lime}{raise}in the\colorbox{lime}{minimum wage help you} \\
        & & & \colorbox{lime}{or anyone you know}over 19\colorbox{lime}{years}   \\
         \cmidrule{2-4}
         &  \multirow{4}{*}{ 2} & \multirow{2}{*} {Ground-truth}  &iv been doing alot more excersise and \\
        & & & eating approx 1200 cals per day\\
        %\cmidrule{2-3}
         \cmidrule{3-4}
      & & \multirow{2}{*}{Recovered} &\colorbox{lime}{iv been doing alot more excersise}and \\
      & & & \colorbox{lime}{eating approx 1200 cals per day} \\
 \bottomrule
    \end{tabular}
\end{table*}

\section{EXTENSION OF MINEGRAD FOR A BATCH OF SAMPLES}

\begin{figure*}%[ht]
\centering
\subfigure[RoBERTa-base]{ \label{fig:RoBERTbase}
  \includegraphics[width=0.23\columnwidth]{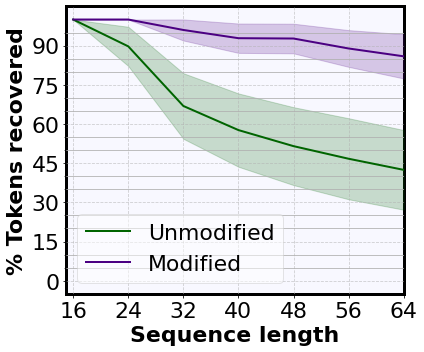} }
  \hspace{-0.3cm}
\subfigure[RoBERTa-large]{ \label{fig:RoBERTlarge}
  \includegraphics[width=0.23\columnwidth]{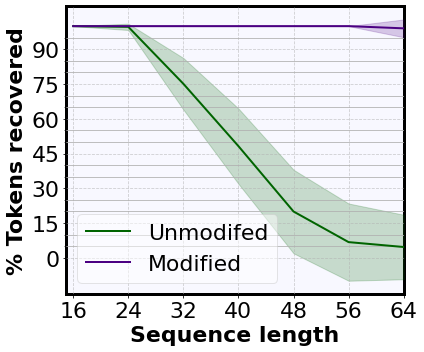} }
  \hspace{-0.3cm}
  \subfigure[BERT-base]{
    \label{fig:BERTbase}
    \includegraphics[width=0.23\columnwidth]{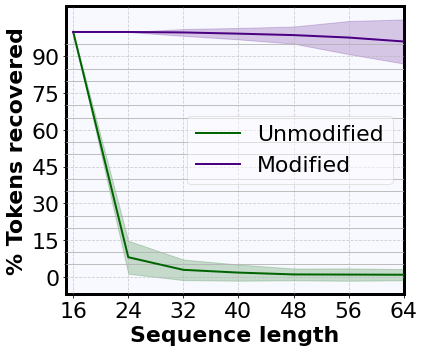}}
\subfigure[BERT-large]{
    \label{fig:BERTlarge}
    \includegraphics[width=0.23\columnwidth]{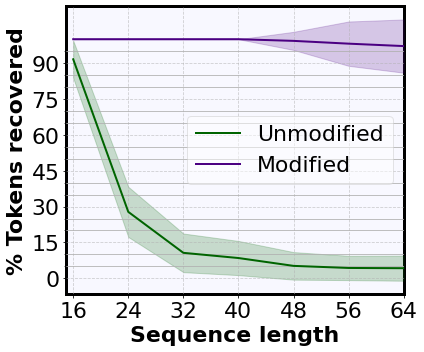}}
\caption{Reconstruction quality for unmodified vs. modified word embeddings (Yelp Review dataset).}
\label{fig:modified_emb}
\end{figure*}

Assume there are $M$ samples in the batch. 
Our attack described in Section \ref{sec:framework} essentially recovers an average of $M$ word embeddings in a target position from the gradient with respect to a particular column of LoRA matrix $\mathbf{B}$. For simplicity, in the following, we keep the description restricted to a target position.
 We denote the word embedding (in the target position) for the $m^{th}$ sample in the batch by $\mathbf{x}(m)$ for $m \in [M]$, and the set of these word embeddings is denoted by $\mathcal{M}$ with $|\mathcal{M}|=M$. 
 From the gradient, the attacker can recover,
\begin{align}
    \mathbf{g} \triangleq \frac{1}{M}\sum_{m =1}^{M}\mathbf{x}(m)
\end{align}
 Now, we design a malicious word embedding layer, where each element is randomly generated from a uniform distribution $\mathcal{U}(-\frac{1}{\sqrt{D}},\frac{1}{\sqrt{D}})$ and $D$ is the embedding dimension as defined in Section 4.

Note that the above distribution has a mean $0$ and variance $\frac{1}{3D}$. Hence, for any word embedding $\mathbf{v}$,
 \begin{align}
     \mathbb{E}[\|\mathbf{v}\|^2] = D \cdot \frac{1}{3D} = \frac{1}{3}
 \end{align}
 Now, for a word embedding $\mathbf{v}$ from the vocabulary where $\mathbf{v} \in \mathcal{M}$, we have
 \begin{align} \label{eq:present}
\mathbb{E}[\mathbf{g}^{\text{T}} \mathbf{v}] &= \frac{1}{M}\mathbb{E}\left[\sum_{m =1}^{M} (\mathbf{x}(m))^{\text{T}} \mathbf{v} \right] \notag\\
&= \frac{1}{M}  \mathbb{E}[\|\mathbf{v}\|^2] + \frac{1}{M}\sum_{\mathbf{x}(m) \neq \mathbf{v}} \mathbb{E}[(\mathbf{x}(m))^{\text{T}} \mathbf{v}] \notag\\
&= \frac{1}{3M}+0=\frac{1}{3M}
 \end{align}
 and for a word embedding $\mathbf{v}$, where $\mathbf{v} \notin \mathcal{M}$, 
 \begin{align} \label{eq:absent}
\mathbb{E}[\mathbf{g}^{\text{T}} \mathbf{v}] = \frac{1}{M}\mathbb{E}\left[\sum_{m =1}^{M} (\mathbf{x}(m))^{\text{T}} \mathbf{v} \right] =0
 \end{align}
 The above properties can be leveraged to find out whether a word embedding from the vocabulary is present in the batch or not. For this, after recovering the average of word embeddings from the gradient, we compute its dot product with the word embeddings in the vocabulary and select the top-$M$ embeddings with highest similarity.  We report the results in Table \ref{tab:token_recovery}, which shows the high success rate of our approach.

\label{App:4}
\section{ADDITIONAL EXPERIMENTS}

\subsection{Modified Word Embedding} \label{app:modified}
In Section \ref{sec:exp}, the experimental results are shown by using the original word embedding layer from the pretrained models. Since this layer is a part of the pretrained model, an adversarial server has the capability to tamper with this layer as well \cite{Feng2024}. The unmodified word  embedding layers from pretrained models contain values with a very small standard deviation (close to $0$). The server can deploy a malicious word embedding layer with a higher standard deviation, leading to a large difference between different word embeddings. This enables the server to distinguish between different word embeddings and recover corresponding words/sub-words with higher accuracy. In Fig. \ref{fig:modified_emb} we demonstrate how an adversarial embedding layer can further enhance the attack performance. For these experiments we consider a  rank $r=4$ and utilize $4$ encoders to recover the first $16$ tokens. We then demonstrate how the recovery of these $16$ tokens is impacted by increasing sequence length. As we observe in Fig. \ref{fig:modified_emb}, for all architectures, performance is significantly improved by leveraging malicious word embeddings over unmodified word embeddings.

\begin{table*} %[ht]
\caption{Average BLEU, ROUGE-L scores and percentage of recovered  tokens from gradient aggregate for $100$ samples.}
    \label{tab:secagg_100samples}
    \centering
    \fontsize{8pt}{9pt}\selectfont
    \begin{tabular}{lcccc}
        \toprule
     \textbf{Sequence length} &   \textbf{Number of users} 
        & BLEU &ROUGE-L  &$\%$ tokens recovered\\
        \midrule
      \multirow{4}{*}{16} &$25$ & $0.99$ &$1.0$  &$100$ \\
        %\cmidrule{2-3}
        &$50$ & $0.99$ &$1.0$  &$100$\\
        &$75$ & $0.99$ &$1.0$  &$100$\\
        &$100$ & $0.99$ &$1.0$  &$100$\\
        \midrule
      \multirow{4}{*}{32} &$25$ & $0.45 \pm 0.27$ &$0.78 \pm 0.12$    &$80.2 \pm 11$\\
        %\cmidrule{2-3}
        &$50$ & $0.45 \pm 0.27$ &$0.78 \pm 0.12$  &$80.2 \pm 11$\\
        &$75$ & $0.45 \pm 0.27$ &$0.78 \pm 0.12$  &$80.2 \pm 11$\\
        &$100$ & $0.45 \pm 0.27$ &$0.78 \pm 0.12$  &$80.2 \pm 11$\\
        \bottomrule
    \end{tabular}
\end{table*}

\subsection{Robustness Against Defense Mechanisms} \label{app:defense}
In this section, we study the robustness of \namespace against existing defense mechanisms such as secure aggregation, pruning, gradient noise and clipping. For this, we run the experiments with the RoBERTa-base pretrained model (unmodified word-embedding layer) and Yahoo Answers dataset with rank $r=4$. We use $4$ encoders to recover $16$ tokens.

\begin{figure}[t]
\centering
\subfigure[BLEU score]{ \label{fig:noise_Bleu}
  \includegraphics[width=0.28\columnwidth]{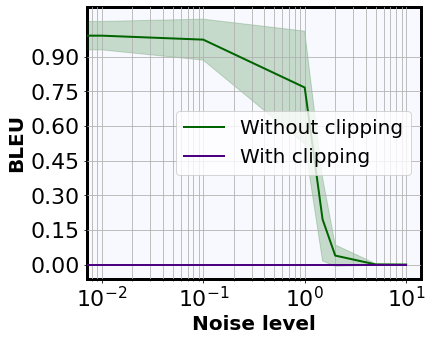} }
  \hspace{-0.3cm}
  \subfigure[Rouge-L score]{
    \label{fig:noise_Rouge}
    \includegraphics[width=0.28\columnwidth]{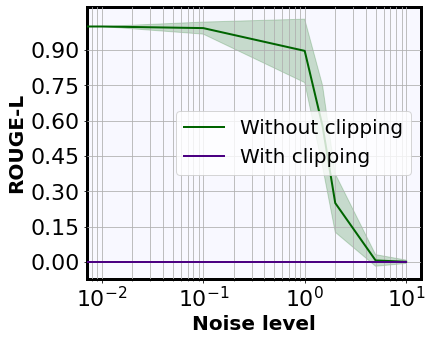}}
\subfigure[Percentage of tokens recovered]{
    \label{fig:noise_acc}
    \includegraphics[width=0.267\columnwidth]{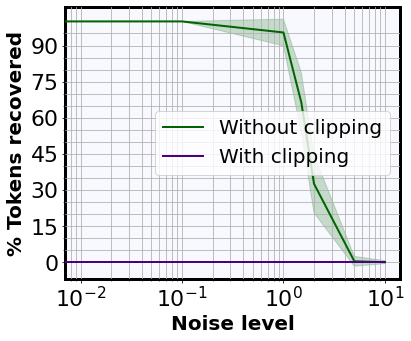}}
\caption{Reconstruction performance under gradient noise and clipping (RoBERTa-base model).}
\label{fig:noise}
\end{figure}

\begin{figure}[t]
\centering
\subfigure[BLEU score]{ \label{fig:pruning_Bleu}
  \includegraphics[width=0.28\columnwidth]{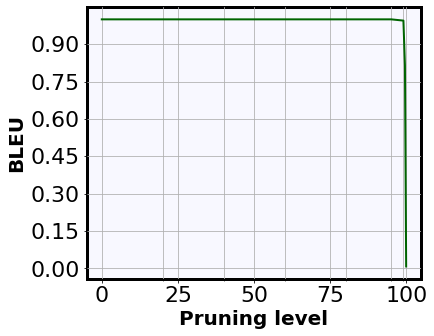} }
  \hspace{-0.3cm}
  \subfigure[Rouge-L score]{
    \label{fig:pruning_Rouge}
    \includegraphics[width=0.28\columnwidth]{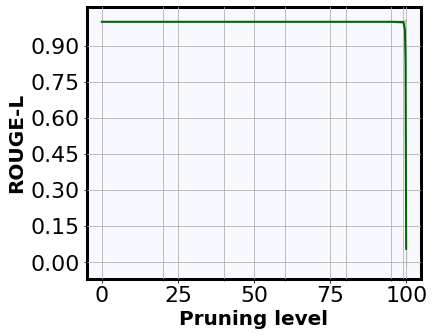}}
\subfigure[Percentage of tokens recovered]{
    \label{fig:pruning_acc}
    \includegraphics[width=0.267\columnwidth]{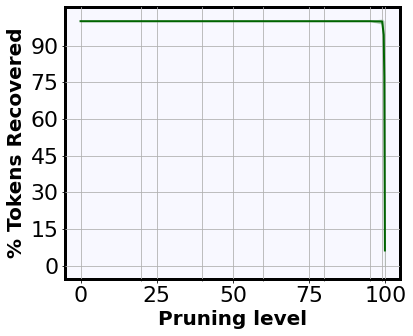}}
\caption{Reconstruction performance under gradient pruning (RoBERTa-base model).}
\label{fig:pruning}
\end{figure}

{\bf Attack to secure aggregation.} We first demonstrate the robustness of \namespace against secure aggregation protocols. Secure aggregation builds on cryptographic primitives \cite{bonawitz2017practical, bell2020secure}, where each user sends a masked gradient to the server. Upon receiving the masked local gradients, the server can decode the true gradient aggregate, but cannot access individual local gradients. Recent works \cite{Pasquini2022, LOKI2024} show how secure aggregation can be bypassed by leveraging model inconsistency, i.e, the server sends different models to different users. However, prior works do not consider the PEFT setup, which is the focus of our work. 
We investigate how individual users can be trapped by \namespace by using model inconsistency under secure aggregation. Let us assume for a target encoder
$t$, 
the server sends malicious LoRA matrices $\mathbf{A}_{O,t}^{u}$, $\mathbf{B}_{O,t}^{u}$ designed according to \eqref{eq:design_A} to a target user $u$, and sends $\mathbf{A}_{O,t}^{v}=\mathbf{0}$, $\mathbf{B}_{O,t}^{v}=\mathbf{0}$ to users $v \neq u$. Then user $u$ would produce the target gradient containing information about target tokens whereas users $v \neq u$ would produce zero gradients. To verify this, we target $2$ users, without loss of generality, user $1$ and user $2$. We set the rank as $r=4$ and leverage encoders $1-4$ for user $1$, and encoders $5-8$ for user $2$ to recover $16$ tokens. For the pretrained model, we use RoBERTa-base.  As we observe in Table \ref{tab:secagg}, even from the gradient aggregate, \namespace can recover fine-tuning texts of these two users. 
% We further aim to recover $32$ tokens by targeting one user in each round and assigning encoders $1-8$ to the target user. As we observe in Table \ref{tab:secagg}, the fine-tuning texts are recovered with high accuracy. 
In Table \ref{tab:secagg_100samples}, we report the average scores across $100$ different recovered samples from aggregated gradients for two different sequence lengths $16$ and $32$. We observe that even with the number of total users increasing, the reconstruction accuracy is not impacted. 

\begin{table}[t]
\centering
 \caption{Attack performance for varying $c_2$ with Yahoo Answers dataset on RoBERTa-base model ($c_1=100$). }
 % \small
\footnotesize
 % \addtolength{\tabcolsep}{-0pt} 
\setlength{\tabcolsep}{3pt}
\begin{tabular}{@{}lc c c c c @{}}\toprule
$c_2$ &$3$ &$5$ &$10$ &$20$ &$30$\\
\midrule\\
 Percentage of tokens recovered & $100$  & $100$ &$100$  &$98.6$  &$39.2$\\
   \bottomrule
\end{tabular} 
\label{table:c_2}
\end{table} 
\begin{table}[htb]
\centering
\caption{Attack performance for varying $c_1$ with Yahoo Answers dataset on RoBERTa-base model ($c_2=3$). }
\footnotesize
\setlength{\tabcolsep}{3pt}
\begin{tabular}{@{}lc c c c  @{}}\toprule
$c_1$ &$5$  &$10$ &$50$ &$100$\\
\midrule\\
 Percentage of tokens recovered & $34.6$ & $98.6$ &$100$  &$100$\\
   \bottomrule
\end{tabular} 
\label{table:c_1}
\end{table}

{\bf Robustness against added noise.} Next, we investigate how well \namespace performs under added noise to gradients \cite{Abadi2016}. In Fig. \ref{fig:noise}, we demonstrate the results with added Gaussian noise with varying levels of standard deviation. As we observe, \namespace successfully retrieves the fine-tuning data with high accuracy even in the presence of noise with standard deviation of up to $1$. We further conduct experiments with gradient clipping, where gradients are clipped to $1$. As we observe, the attack is unsuccessful under clipping, suggesting clipping could be a better defense. %Hence, clipping can be used as a potential defense, however its impact on overall model utility needs to be studied in-depth.

{\bf Robustness against pruning.} We now demonstrate how \namespace attack is affected by gradient pruning \cite{Yujun2018, Alistarh2018}, where gradient elements with magnitude below a certain percentage are set to $0$. As we observe in Fig. \ref{fig:pruning}, even for up to $99 \%$ pruning, reconstruction remains unaffected. This can be attributed to the fact that only the target gradients (gradients from LoRA modules of either value weights or output projection weights) are non-zero. Hence, mostly non-target zero gradients are pruned out, while retaining the target gradients.

\subsection{Impact of Varying $c_1$ and $c_2$}

In Table \ref{table:c_2}, we report the attack performance with varying $c_2$ while keeping $c_1$ fixed. As we see, attack success degrades with higher $c_2$.
Moreover, under a fixed $c_2$, choosing a lower value for $c_1$ causes an approximation error in Eqs. (6) and (7), resulting in degraded attack performance as evident in Table \ref{table:c_1}.

\label{App:5}

 \subsection{Decoder-Based Architecture}\label{app:GPT2} 
 \begin{table}[t]
\caption{Mean and standard deviation of the performance metrics  across $100$ samples with sequence length $16$ for different datasets (GPT-2). }
  %Lower LPIPS/higher SSIM implies better reconstruction.  } 
\label{table:GPT2}
\centering
\fontsize{8pt}{9pt}\selectfont
%\footnotesize
 % \addtolength{\tabcolsep}{-0pt} 
\setlength{\tabcolsep}{3pt}
\begin{tabular}{@{}lc c c c @{}}\toprule
{\bf Dataset} &\multicolumn{3}{c}{\bf Metrics}\\
\midrule\\
& {\bf ROUGE-L} & {\bf BLEU}  & {\bf $\%$ Tokens Recovered}   \\
\cmidrule{2-4}\\
% \cmidrule(lr){2-4} \cmidrule(lr){5-7}
% \midrule 
AG's News &$0.99 \pm 0.04  $ &$0.96 \pm 0.02$  &$99.7 \pm 1$\\
DBPedia &$0.94 \pm 0.15  $ &$0.98 \pm 0.03$  &$99.2 \pm 2$\\
Yahoo Answers &$0.996 \pm 0.02  $ &$0.98 \pm 0.08$  &$99.3 \pm 2$\\
Yelp Review &$0.996 \pm 0.02  $ &$0.99 \pm 0.05$  &$99.6 \pm 2$ \\
TREC &$0.998 \pm 0.01  $ &$0.994 \pm 0.03$  &$99.7 \pm 1$ \\
   \bottomrule
\end{tabular} 
% \addtolength{\tabcolsep}{2pt}

 %  \vspace{0.1cm}
  
 % \vspace{0.3cm}
\end{table}

 We now demonstrate the attack performance of \namespace with GPT-2 \cite{Radford2019} as the pretrained model. Unlike encoder-based architectures like BERT or RoBERTa, GPT-2 follows a
unidirectional self-attention mechanism. For classification tasks, either a class token is added to the end of all other tokens,
%right, 
% {\color{orange} Need to clarify what we mean by "right". }
or
the last token is used for classification. Same design principles
as described in Section \ref{sec:framework} can also be utilized for GPT-2. In Table \ref{table:GPT2}, we report the mean and standard deviation across $100$ samples for different metrics. As we observe, the tokens are recovered with high success rate. 
%We report our results in Table II in https://rb.gy/pwlg2p, where
%we provide the mean and standard deviation across 100 samples
%for different performance metrics.

\begin{figure}[t]
  \centering 
   \subfigure[Original]{\label{fig:original_CIFAR10}
    \includegraphics[width=0.5\linewidth]{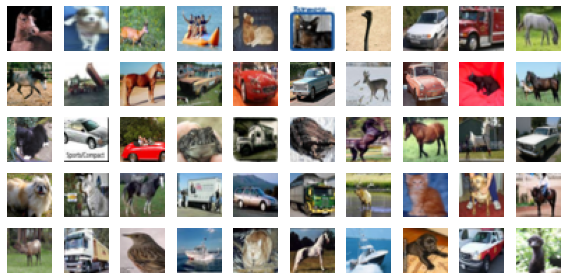} 
    }   
    \hspace{0.5cm}
  \subfigure[Recovered]{\label{fig:rec_CIFAR10}
    \includegraphics[width=0.5\linewidth]{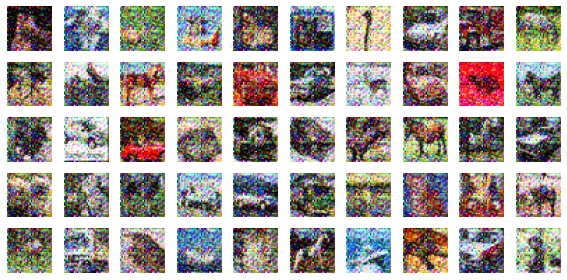} 
    }
\vspace{-0.2cm}  \caption{Reconstruction performance of MineGrad with $r=4$ on CIFAR-10. }
  \label{fig:recovery_CIFAR10}
%\vspace{-0.3cm}
\end{figure}

\begin{table} %[ht]
\caption{Recovered tokens for multiple datasets with both modified and unmodified word embeddings (rank $r=4$). We leverage fine-tuning gradients from $8$ encoders to recover $32$ tokens.}
    \label{tab:long}
\footnotesize
    \centering
    \fontsize{8pt}{9pt}\selectfont
    \begin{tabular}{lcc}
        \toprule
        \textbf{Dataset} & \multicolumn{2}{c}{\textbf{Fine-tuning sentences}}    \\
        \midrule
        \multirow{6}{*}{AG's} & \multirow{2}{*}{Ground-truth text} &Google gets a bounce, ends its first day up 18 percent   \\
        & &Shares of Google leaped \$15.34, or 18 percent, to \$100. \\
        \cmidrule{2-3}
        & \multirow{2}{*}{Recovered (unmodified)} &\colorbox{lime}{Google} gets a \colorbox{lime}{bounce}, \colorbox{lime}{ends its first day up 18 percent}   \\
        & &\colorbox{lime}{Shares} of \colorbox{lime}{Google leaped \$}15.34, \colorbox{lime}{or 18 percent}, to \colorbox{lime}{\$}100. \\
        \cmidrule{2-3}
        & \multirow{2}{*}{Recovered (modified)} &\colorbox{lime}{Google gets a bounce}, \colorbox{lime}{ends its first day up 18 percent}   \\
        & &\colorbox{lime}{Shares of} \colorbox{lime}{Google leaped \$15}.\colorbox{lime}{34}, \colorbox{lime}{or 18 percent}, \colorbox{lime}{to \$100}. \\ 
         \midrule
        \multirow{9}{*}{DBPedia} & \multirow{3}{*}{Ground-truth text} &Takuma Kanaiwa is a New York City based musician and \\
        & & electrical engineer. His music is known for transcending \\
        & & genres such as free jazz and Japanese folk. \\
        \cmidrule{2-3}
        & \multirow{3}{*}{Recovered (unmodified)} &\colorbox{lime}{Takuma Kanaiwa} is a \colorbox{lime}{New York City based musician} and \\
        & & \colorbox{lime}{electrical engineer}. His\colorbox{lime}{music} is \colorbox{lime}{known} for \colorbox{lime}{transcending} \\
        & & \colorbox{lime}{genres such} as \colorbox{lime}{free jazz} and \colorbox{lime}{Japanese folk}. \\
        
        \cmidrule{2-3}
        & \multirow{3}{*}{Recovered (modified)} &\colorbox{lime}{Takuma Kanaiwa is a New York City based musician and} \\
        & & \colorbox{lime}{electrical engineer}. \colorbox{lime}{His music is known for transcending} \\ 
        & & \colorbox{lime}{genres such as free jazz and Japanese folk}. \\
        \midrule
        \multirow{9}{*}{Yahoo} & \multirow{3}{*}{Ground-truth text} &In the san francisco bay area, does it make sense \\
        & & to rent or buy ? the prices of rent and the price \\
        & & of buying does not make sense to me \\
        \cmidrule{2-3}
        & \multirow{3}{*}{Recovered (unmodified)} &In the \colorbox{lime}{san francisco bay area}, does it make \colorbox{lime}{sense} \\
        & & to\colorbox{lime}{rent}or buy\colorbox{lime}{?}the \colorbox{lime}{prices} of \colorbox{lime}{rent} and the \colorbox{lime}{price} \\ 
        & & of buying does\colorbox{lime}{not}make\colorbox{lime}{sense}to\colorbox{lime}{me} \\
        
        \cmidrule{2-3}
        & \multirow{3}{*}{Recovered (modified)} &\colorbox{lime}{In the san francisco bay area}, \colorbox{lime}{does it make sense} \\
        & & \colorbox{lime}{to rent or buy ? the prices of rent and the price} \\ 
        & & \colorbox{lime}{of buying does not make sense to me} \\
        \midrule
        \multirow{9}{*}{Yelp} & \multirow{3}{*}{Ground-truth text} &I try to visit every time I'm in LV because I like \\
        & &the food here but I'm only giving 3 stars because \\
        & &of the controversy around their calorie counts. \\
        \cmidrule{2-3}
        & \multirow{3}{*}{Recovered (unmodified)} &I \colorbox{lime}{try} to \colorbox{lime}{visit every} time I\colorbox{lime}{'m} in \colorbox{lime}{LV because} I \colorbox{lime}{like} \\
        & & the \colorbox{lime}{food}\colorbox{lime}{here} but I\colorbox{lime}{'m only giving 3 stars because} \\
        & & of the \colorbox{lime}{controversy around} their \colorbox{lime}{calorie counts}. \\
        
        \cmidrule{2-3}
        & \multirow{3}{*}{Recovered (modified)} &\colorbox{lime}{I try to visit every time I'm in LV because} \\
        & & \colorbox{lime}{I like the food}\colorbox{lime}{here but I'm only giving 3 stars because} \\
        & & \colorbox{lime}{of the controversy around their calorie counts}. \\
        \midrule
        \multirow{9}{*}{TREC} & \multirow{2}{*}{Ground-truth text} &Who was the author of the book about computer hackers \\
        & & called `` The Cuckoo 's  Egg : Tracking a Spy  \\
        & & Through the Maze of Computer Espionage '' ? \\
        \cmidrule{2-3}
        & \multirow{3}{*}{Recovered (unmodified)} &Who was the \colorbox{lime}{author} of the \colorbox{lime}{book about computer hackers} \\
        & & \colorbox{lime}{called ``} The \colorbox{lime}{Cuckoo '}s\colorbox{lime}{Egg : Tracking} a \colorbox{lime}{Spy} \\
        & & \colorbox{lime}{Through} the \colorbox{lime}{Maze} of \colorbox{lime}{Computer Espionage ''} ? \\
        
        \cmidrule{2-3}
        & \multirow{3}{*}{Recovered (modified)} &\colorbox{lime}{Who was the author of the book about computer hackers} \\
        & & \colorbox{lime}{called `` The Cuckoo 's}\colorbox{lime}{Egg : Tracking a Spy} \\ 
        & & \colorbox{lime}{Through the Maze of Computer Espionage ''}? \\
        %\midrule
         %\midrule
        \bottomrule
    \end{tabular}
\end{table}

\subsection{Recovered Images from CIFAR-10 Dataset} \label{app:img}

In Fig. \ref{fig:recovery_CIFAR10}, we demonstrate the reconstruction of sample images from CIFAR-10 dataset. As we observe, images are recovered with high fidelity.

\subsection{More Examples of Recovered Sequences} \label{app:long_seq}
In Table \ref{tab:long}, we show reconstruction results for all $32$ tokens from sequences with length $32$. For this, we leverage gradients from $8$ encoders with rank $r=4$.

\label{App:3}

\subsection{Trainable LayerNorm/Embeddings}
\begin{table*}[h]
 \caption{Average Scores in terms of BLEU and Percentage of tokens recovered (TokRec) over $100$ samples with trainable LayerNorm and word-embeddings.}
 \vspace{-0.1cm}
 \label{tab:100_samples_LN}
 \centering
 \footnotesize
 \begin{tabular}{lcccccccc}
    \toprule
    \textbf{Dataset} &
    \multicolumn{2}{c}{\textbf{RoBERTa-base}} &
    \multicolumn{2}{c}{\textbf{RoBERTa-large}} &
    \multicolumn{2}{c}{\textbf{BERT-base}} &
    \multicolumn{2}{c}{\textbf{BERT-large}} \\
    
    \midrule
    & BLEU & \%TokRec
    & BLEU & \%TokRec
    & BLEU & \%TokRec
    & BLEU & \%TokRec \\
    \midrule
    Yahoo    & $1.0$ & $100$ & $1.0$ & $100$ & $1.0$ & $100$ & $0.81$ & $95$ \\
    Yelp     & $0.98$ & $100$ & $0.98$ & $100$ & $0.98$ & $100$ & $0.73$ & $93$ \\
    DBPedia  & $0.99$ & $100$ & $0.99$ & $100$ & $1.0$ & $100$ & $0.79$ & $93$ \\
    TREC     & $1.0$ & $100$ & $1.0$ & $100$ & $0.98$ & $100$ & $0.82$ & $94$ \\
    AG's     & $1.0$ & $100$ & $1.0$ & $100$ & $1.0$ & $100$ & $0.82$ & $94$ \\
    
    \bottomrule
 \end{tabular}
\end{table*}

Our attack remains successful when LayerNorm/word embeddings are  trainable. For LayerNorm, \eqref{eq:p} guarantees that the target token embeddings reach the next encoder’s LoRA modules without distortion as in \eqref{eq:MSAop}. Hence, target gradients with respect to the LoRA matrix $\mathbf{B}$ remain exactly as in \eqref{eq:chain_rule} and still reveal the fine-tuning tokens - additional gradients with respect to LayerNorm do not affect this signal. A similar argument also holds for word embeddings. Table \ref{tab:100_samples_LN} demonstrates our results for reconstruction performance, where we report the BLEU Scores, as well as the percentage of tokens recovered for sequence length $16$ and $r=4$. \label{App:trainable}
\subsection{Different Optimizers}
\begin{table*}[h!]
\centering
\caption{Average scores across $100$ samples from Yahoo  Answers Dataset with different optimizers (RoBERTa-base). 
}
\label{tab:optimizers}
\begin{tabular}{lcccc}
\hline
\textbf{Seq. Length} & \textbf{Optimizer} & \textbf{Rouge-L} & \textbf{BLEU} & \textbf{\% Tokens Recovered} \\
\hline
\multirow{3}{*}{16} 
& SGD     & 1.00 & 1.00 & 100 \\
& Adam    & 1.00 & 1.00 & 100 \\
& AdaGrad & 1.00 & 1.00 & 100 \\
\hline
\multirow{3}{*}{32} 
& SGD     & 0.77 & 0.45 & 79 \\
& Adam    & 0.76 & 0.43 & 78 \\
& AdaGrad & 0.77 & 0.44 & 79 \\
\hline
\end{tabular}
\end{table*}

Our original framework and derivations consider conventional SGD as the optimizer, as was typically considered also in prior attacks \cite{Hong2023, Fowl2022, Fowl2023, Petrov2024, LOKI2024}, but adaptive optimizers such as Adam or AdaGrad can also be considered as follows. For server side optimizers \cite{wang2022communication}, note that an adversarial server can bypass such optimizers, so the attack is unchanged. 
For user-side optimization \cite{wu2023faster},  since first and second moments of Adam are initialized as zero, after one local step and bias correction, the model update becomes,
\begin{align}
 \mathbf{\hat{g}}_u =  \eta\frac{\mathbf{g}_u}{|\mathbf{g}_u|+\epsilon}
\end{align} 
where $\mathbf{g}_u$  is the gradient of user $u$, $\eta$ is the step-size, and $\epsilon$ is a small parameter  ($\sim10^{-8}$). Note that $\mathbf{\hat{g}}_u$ preserves the sign pattern of $\mathbf{g}_u$. Since the target gradient in \eqref{eq:chain_rule} is a scaled token embedding,  cosine similarity still recovers the correct embedding and attack remains effective. The same also holds for AdaGrad. Because the attack is one-shot, the adversary only needs the first round. Table \ref{tab:optimizers} reports our results with different optimizers.\label{App:optimizers}
\subsection{Label Smoothing}
\begin{table*}[h!]
\centering
\caption{Average scores across $100$ samples from Yahoo Answers Dataset with different levels of smoothing (RoBERTa-base).}
\label{tab:label_smoothing}
\begin{tabular}{cccccc}
\hline
\textbf{Seq. Length} & \textbf{$\alpha$} & \textbf{Rouge-L} & \textbf{BLEU} & \textbf{\% Tokens Recovered } \\
\hline
\multirow{4}{*}{16}
& 0.1 & 1.00 & 1.00 & 100 \\
& 0.2 & 1.00 & 1.00 & 100 \\
& 0.3 & 1.00 & 1.00 & 100 \\
& 0.4 & 1.00 & 1.00 & 100 \\
\hline
\multirow{4}{*}{32}
& 0.1 & 0.76 & 0.41 & 78 \\
& 0.2 & 0.74 & 0.38 & 77 \\
& 0.3 & 0.74 & 0.37 & 76 \\
& 0.4 & 0.73 & 0.36 & 76 \\
\hline
\end{tabular}
\end{table*}

We next demonstrate attack performance under label smoothing. 
In Table \ref{tab:label_smoothing}, we replace each label $i \in [C]$ with,
\begin{align}
    l_i=(1-\alpha)\delta_{i}+\frac{\alpha}{C} 
\end{align}
where $\delta_{i}$ is the Dirac delta, $C$ is the number of classes, and $\alpha$ is the smoothing parameter \cite{szegedy2016rethinking}. In \eqref{eq:chain_rule} one-hot labels result in $z_1 - l_1 = 1$. On the other hand, under smoothing, $l_1 = \alpha/C$, hence $z_1-l_1 = 1-\alpha/C$, which stays close to $1$ for typical $\alpha$ choices, e.g., $0.1$ \cite{szegedy2016rethinking}. 
Thus, embeddings remain almost undistorted, and attack succeeds. \label{App:label_smoothing}
\subsection{Dropout Rate}
\begin{table*}[h!]
\centering
\caption{Average scores across 100 samples for different dropout rates (RoBERTa-base, sequence length 16).}
\label{tab:dropout_tokrec}
\begin{tabular}{l c c c c} % <-- 4 columns now
\toprule
\textbf{Dataset} & \textbf{Dropout rate ($p$)} & \textbf{Rouge-L} & \textbf{BLEU} & \textbf{\% Tokens Recovered} \\
\midrule
\multirow{3}{*}{Yahoo}
 & 0.1 & 0.91 & 0.77 & 94.46 \\
 & 0.2 & 0.79 & 0.52 & 82.68 \\
 & 0.4 & 0.52 & 0.20 & 59.35 \\
\midrule
\multirow{3}{*}{Yelp}
 & 0.1 & 0.90 & 0.72 & 92.58 \\
 & 0.2 & 0.76 & 0.44 & 80.29 \\
 & 0.4 & 0.60 & 0.21 & 63.42 \\
\midrule
\multirow{3}{*}{DBpedia}
 & 0.1 & 0.90 & 0.74 & 92.17 \\
 & 0.2 & 0.81 & 0.54 & 83.00 \\
 & 0.4 & 0.56 & 0.25 & 60.50 \\
\midrule
\multirow{3}{*}{TREC}
 & 0.1 & 0.91 & 0.79 & 95.28 \\
 & 0.2 & 0.78 & 0.48 & 81.60 \\
 & 0.4 & 0.49 & 0.16 & 58.03 \\
\midrule
\multirow{3}{*}{AG News}
 & 0.1 & 0.91 & 0.78 & 94.46 \\
 & 0.2 & 0.79 & 0.50 & 83.83 \\
 & 0.4 & 0.49 & 0.16 & 58.63 \\
\bottomrule
\end{tabular}
\end{table*}

We evaluate our attack under DropKey \cite{li2023dropkey}, which applies dropout to the scaled dot-product scores in the attention layer, by adding a dropout mask before softmax, where each element  is sampled as, 
\begin{align}
    x = \left\{\begin{matrix} 
 0&\text{ with prob. } 1-p \\
 -\infty &\text{ with prob. } p
 \end{matrix} \right. \vspace{-0.2cm}
\end{align} 
where the dropout rate is $p\in [0,1]$. 
Table \ref{tab:dropout_tokrec} reports the reconstruction performance for different dropout rates $p$. As expected, lower dropout yields higher success. \label{App:dropout}

%%%%%%%%%%%%%%%%%%%%%%%%%%%%%%%%%%%%%%%%%%%%%%%%%%%%%%%%%%%%%%%%%%%%%%%%%%%%%%%
%%%%%%%%%%%%%%%%%%%%%%%%%%%%%%%%%%%%%%%%%%%%%%%%%%%%%%%%%%%%%%%%%%%%%%%%%%%%%%%

\end{document}